\documentclass[letterpaper]{article} 
\usepackage{aaai2026}   
\usepackage{times}  
\usepackage{helvet}  
\usepackage{courier}  
\usepackage[hyphens]{url}  
\usepackage{graphicx} 
\usepackage{natbib}  
\usepackage{caption} 
\usepackage{algorithm}
\usepackage{algorithmic}
\usepackage{subcaption}
\usepackage{caption}
\usepackage{graphicx}

\usepackage{newfloat}
\usepackage{listings}
\DeclareCaptionStyle{ruled}{labelfont=normalfont,labelsep=colon,strut=off} 
\floatstyle{ruled}
\newfloat{listing}{tb}{lst}{}
\floatname{listing}{Listing}
\usepackage{booktabs}
\usepackage{amsmath}
\usepackage{amssymb}

\title{Overconfidence in LLM-as-a-Judge: Diagnosis and Confidence-Driven Solution}
\author{
    Zailong Tian\equalcontrib\textsuperscript{\rm 1}, 
    Zhuoheng Han\equalcontrib\textsuperscript{\rm 2}, 
    Yanzhe Chen\textsuperscript{\rm 4}, 
    Haozhe Xu\textsuperscript{\rm 2}, 
    Xi Yang\textsuperscript{\rm 3}, 
    Richeng Xuan\textsuperscript{\rm 3}\thanks{Corresponding author.}, 
    Houfeng Wang\textsuperscript{\rm 2}\thanks{Corresponding author.}, 
    Lizi Liao\textsuperscript{\rm 1}\thanks{Corresponding author.} 
}
\affiliations{
    \textsuperscript{\rm 1}School of Computing and Information Systems, Singapore Management University\\
    \textsuperscript{\rm 2}State Key Laboratory for Multimedia Information Processing, Peking University\\
    \textsuperscript{\rm 3}Beijing Academy of Artificial Intelligence\\
    \textsuperscript{\rm 4}School of Computing, National University of Singapore\\
    \{zltian,lzliao\}@smu.edu.sg,
    \{2100017789,xhzgenius\}@stu.pku.edu.cn,
    chenyanzhe@u.nus.edu,
    \{rcxuan,yangxi\}@baai.ac.cn,
    wanghf@pku.edu.cn
}

\usepackage{bibentry}
\begin{document}
\maketitle
\begin{abstract}
Large Language Models (LLMs) are widely used as automated judges, where practical value depends on both accuracy and trustworthy, risk-aware judgments.
Existing approaches predominantly focus on accuracy, overlooking the necessity of well-calibrated confidence, which is vital for adaptive and reliable evaluation pipelines. 
In this work, we advocate a shift from accuracy-centric evaluation to confidence-driven, risk-aware LLM-as-a-Judge systems, emphasizing the necessity of well-calibrated confidence for trustworthy and adaptive evaluation.
We systematically identify the \textbf{Overconfidence Phenomenon} in current LLM-as-a-Judges, where predicted confidence significantly overstates actual correctness, undermining reliability in practical deployment. 
To quantify this phenomenon, we introduce \textbf{TH-Score}, a novel metric measuring confidence-accuracy alignment. 
Furthermore, we propose \textbf{LLM-as-a-Fuser}, an ensemble framework that transforms LLMs into reliable, risk-aware evaluators. Extensive experiments demonstrate that our approach substantially improves calibration and enables adaptive, confidence-driven evaluation pipelines, achieving superior reliability and accuracy compared to existing baselines. 

\end{abstract}
\section{Introduction}

The widespread adoption of large language models (LLMs) as automated judges—termed the LLM-as-a-Judge paradigm—has revolutionized the evaluation of AI-generated content by offering scalability and efficiency over traditional human annotation~\cite{zheng2023judging}. In this paradigm, LLMs act as evaluators, with one common application being pairwise comparisons where the model decides which of two text segments is better based on criteria like quality, relevance, or coherence. However, the practical value of these systems depends not only on accuracy but also on trustworthy, risk-aware judgments that can adapt to real-world deployment scenarios. Existing approaches, such as FairEval~\cite{wang2023FairEval} and JudgeBench~\cite{tan2024judgebench}, predominantly emphasize accuracy, often overlooking the critical role of well-calibrated confidence. This calibration, defined as the alignment between a model's predicted confidence and its actual correctness, is essential for building adaptive evaluation pipelines. For instance, well-calibrated confidence allows high-confidence outputs to be automatically accepted, minimizing manual intervention, while low-confidence cases can be flagged for human review~\cite{li2024llms}. In this work, we advocate a fundamental shift from accuracy-centric evaluations to confidence-driven, risk-aware LLM-as-a-Judge systems, prioritizing calibration to ensure reliable and trustworthy assessments.

\begin{figure}[t!]
    \centering
    \includegraphics[width=1\linewidth]{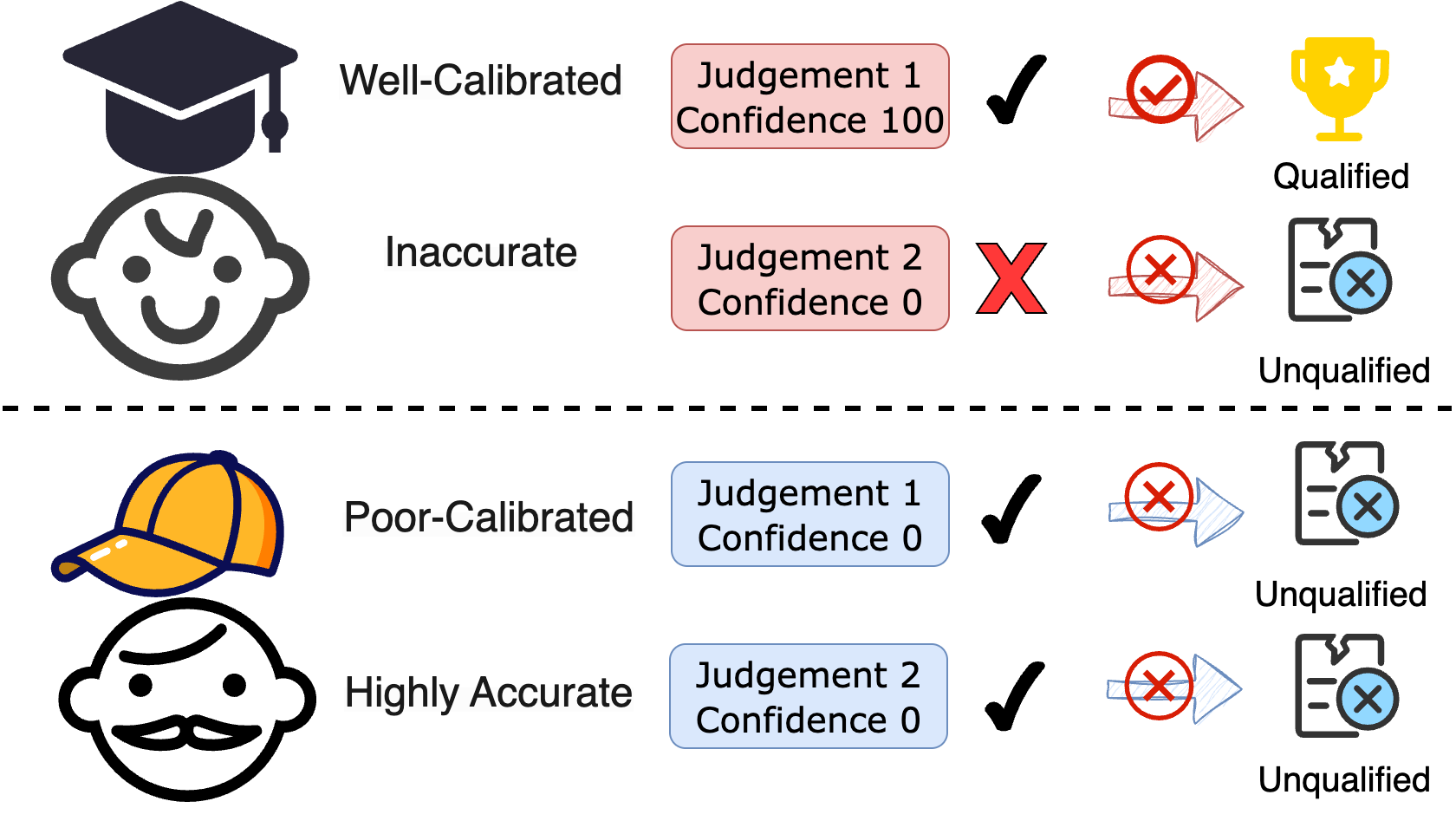}
    \caption{Well-calibrated judgments align confidence with accuracy, qualifying correct high-confidence predictions ($\checkmark \rightarrow$ qualified) and disqualifying inaccurate low-confidence ones ($\times \rightarrow$ unqualified). In contrast, poorly calibrated models mismatch confidence and accuracy, often disqualifying even highly accurate judgments ($\checkmark \rightarrow$ unqualified), leading to unreliable outcomes overall.}
\end{figure}

Despite these potential benefits, current LLM-as-a-Judge systems suffer from a pervasive Overconfidence Phenomenon, where predicted confidence levels significantly overstate actual correctness~\cite{mielke2022reducing,zhou2023navigating}, thereby undermining reliability in practical applications. Through systematic analysis, we observe that state-of-the-art LLMs exhibit this issue prominently, leading to inflated confidence scores that do not reflect true performance~\cite{zhao2021calibrate}. This misalignment results in substantial risks: overconfident models may propagate erroneous judgments without detection, eroding the efficiency gains of automated evaluation, while also complicating downstream decision-making in pipelines~\cite{gu2024survey}. Furthermore, existing benchmarks and metrics exacerbate the problem by focusing on aggregate accuracy without addressing confidence alignment, introducing biases such as response length or model familiarity that distort calibration assessments~\cite{chen-etal-2024-humans,zheng2023judging,wang2023FairEval}. Consequently, the lack of calibration-aware tools limits the deployment of LLMs as dependable evaluators in high-stakes environments.

To address these challenges, we introduce TH-Score, a novel metric that quantifies confidence-accuracy alignment by focusing on critical high- and low-confidence intervals, where practical decisions hinge. Unlike traditional metrics like accuracy or Expected Calibration Error (ECE)—which ignore confidence or overlook key thresholds—TH-Score balances accuracy within these intervals against their coverage, rewarding aligned successes (e.g., high-confidence correct predictions) while penalizing mismatches like overconfident errors. This makes TH-Score a principled tool for detecting the Overconfidence Phenomenon under LLM-as-a-Judge scenario, highlighting cases where high confidence fails to match actual correctness.

Furthermore, we propose LLM-as-a-Fuser, an ensemble framework that leverages a dedicated "fuser" LLM to synthesize judgments and critiques from multiple models, transforming LLMs into reliable, risk-aware evaluators. By integrating diverse perspectives, LLM-as-a-Fuser significantly enhances calibration. Extensive experiments on a widely-used benchmark demonstrate that our approach achieves superior calibration, reliability, and overall accuracy compared to existing baselines, paving the way for more trustworthy LLM-as-a-Judge systems in practical settings.

In a nutshell, our contributions are threefold:
\begin{itemize}
    \item \textbf{Overconfidence Phenomenon}: We systematically identify and characterize the overconfidence in LLM-as-a-Judge, where confidence overstates correctness, limiting risk-aware evaluation.
    \item \textbf{Metric Innovation}: We introduce TH-Score, a novel metric quantifying confidence-accuracy alignment for trustworthy LLM judgments.
    \item \textbf{Framework Advancement}: We propose LLM-as-a-Fuser, an ensemble approach that enhances calibration, enabling adaptive, confidence-driven pipelines with superior reliability and accuracy.
\end{itemize}

\section{Related Work}

\subsection{LLM-as-a-Judge}
LLMs are increasingly used as automated evaluators for text quality. \cite{zheng2023judging} showed GPT-4 aligns with human judgments in pairwise comparisons, but proprietary APIs limit reproducibility. PandaLM~\cite{wang2023pandalm} introduced a 7B-parameter local evaluator with 94\% agreement with ChatGPT, supporting offline use. JudgeLM~\cite{zhu2025judgelm} and Agent-as-a-Judge~\cite{zhuge2024agent} use modular frameworks with memory and planning, cutting DevAI evaluation costs by 97\%. However, alignment between model confidence and accuracy is often ignored, causing inconsistent judgments. Meta's self-rewarding models~\cite{wu2024meta} generate and evaluate outputs iteratively, but calibration needs further study.

As LLM-based judges gain traction for evaluating and enhancing LLMs, various benchmarks have emerged to gauge their effectiveness. Prior works like LLMEval ~\cite{lin2023llmevalunifiedmultidimensionalautomatic}, MTBench, and FairEval primarily assess how well LLM-based judges align with subjective human preferences, often emphasizing stylistic differences over factual and logical accuracy. Similarly, LLMBar ~\cite{zeng2024evaluatinglargelanguagemodels} evaluates judges based on their ability to follow instructions, using response pairs with clear ground truth labels tied to instruction adherence. In contrast, JudgeBench offers a novel benchmark specifically designed to test LLM-based judges’ reasoning capabilities. It features 350 challenging response pairs across knowledge, reasoning, math, and coding domains, each containing one objectively correct and one subtly incorrect response, prioritizing factual and logical correctness over subjective or stylistic factors.

\subsection{Calibration in LLMs}
Accurate calibration, aligning a model's confidence with its accuracy, is crucial for reliable LLM applications. Traditional methods like temperature scaling~\cite{guo2017calibration} adjust confidence with a single scalar but are less effective for large models, while Bayesian methods are computationally infeasible. Recent approaches, such as the Thermometer method~\cite{shen2024thermometer}, train auxiliary models for recalibration, achieving top uncertainty quantification across 12 benchmarks, and SPACE~\cite{yi2024generation} uses lightweight linear layers for dynamic confidence adjustment. However, most techniques focus on single models, missing multi-model aggregation benefits, and Collaborative Calibration~\cite{yang2024confidence} reduces overconfidence via multi-agent deliberation but requires significant resources. Current research lacks focus on calibration's impact on downstream tasks like data generation, where confidence filtering affects output quality, warranting further exploration.

\subsection{Uncertainty Quantification and Reward Modeling}
Uncertainty-aware frameworks bridge calibration and practical applications. Generating with Confidence~\cite{lin2023generating} combines Monte Carlo dropout and response length analysis to filter low-confidence outputs, demonstrating that well-calibrated models yield higher-quality synthetic data. Inference-Time Scaling~\cite{liu2025inference} dynamically aligns reward models with human preferences, indirectly improving calibration through gradient-free optimization. However, these approaches often assume static datasets, failing to address the iterative nature of LLM-as-a-Judge workflows. Benchmarking LLMs via Uncertainty Quantification~\cite{ye2024benchmarking} reveals that calibration degrades under distribution shifts (e.g., domain-specific tasks), underscoring the need for adaptive methods.

\section{Overconfidence in LLM-as-a-Judge}\label{sec:Overconfidence}
In the LLM-as-a-Judge paradigm, models are typically required to select the superior option from pairwise samples. However, the reliability of model predictions warrants careful examination, particularly regarding the Overconfidence Phenomenon—a tendency for language models to display predicted confidence levels that significantly exceed their actual accuracy, resulting in calibration gaps that undermine reliability. Underconfident models tend to underestimate their own accuracy, while overconfident ones overestimate their judgment correctness. Such biases introduce noisy signals that can adversely affect the performance of downstream tasks (e.g., reward modeling). Particularly in unsupervised or weakly-supervised scenarios, developing a well-calibrated model where judgment capability aligns with confidence becomes crucial. By acquiring confidence of model judgments, we can not only filter out low-accuracy predictions but also effectively identify high-accuracy decisions, thereby enhancing the overall system reliability.

\subsection{How to measure confidence in LLMs? }
We employed three methods for calculating confidence: Self-Confidence (SC), Multiple-Prompting (MP) confidence, and Logprob confidence.

\noindent \textbf{SC setting}: We prompt the model to output both the result and its confidence. Model's temperature is set to 0 to ensure the reproducibility of the setting.

\noindent \textbf{MP setting}: We adopt a method similar to SimpleQA~\cite{simpleqa}, but reduce the number of requests from 100 to 10 for efficiency, while keeping the temperature at 0.7. The final reply is determined by majority voting, and the confidence is the count of the chosen response over 10.

\noindent \textbf{LogP Setting.} In this setting, confidence scores are derived from softmax-normalized logits for the final output tokens (e.g., `A' or `B'). For a binary choice task with options $A$ and $B$, and corresponding logits $l_A$ and $l_B$, we first compute the softmax probabilities:
\[
p(A) = \frac{e^{l_A}}{e^{l_A} + e^{l_B}}, \quad p(B) = \frac{e^{l_B}}{e^{l_A} + e^{l_B}}.
\]
Confidence is then defined as the maximum probability:
\[
\text{Confidence}_{\text{logp}} = \max\left(p(A), p(B)\right).
\]
Temperature is set to 0 to ensure deterministic outputs.

\begin{figure*}[ht!]
    \centering
\includegraphics[width=0.96\textwidth]{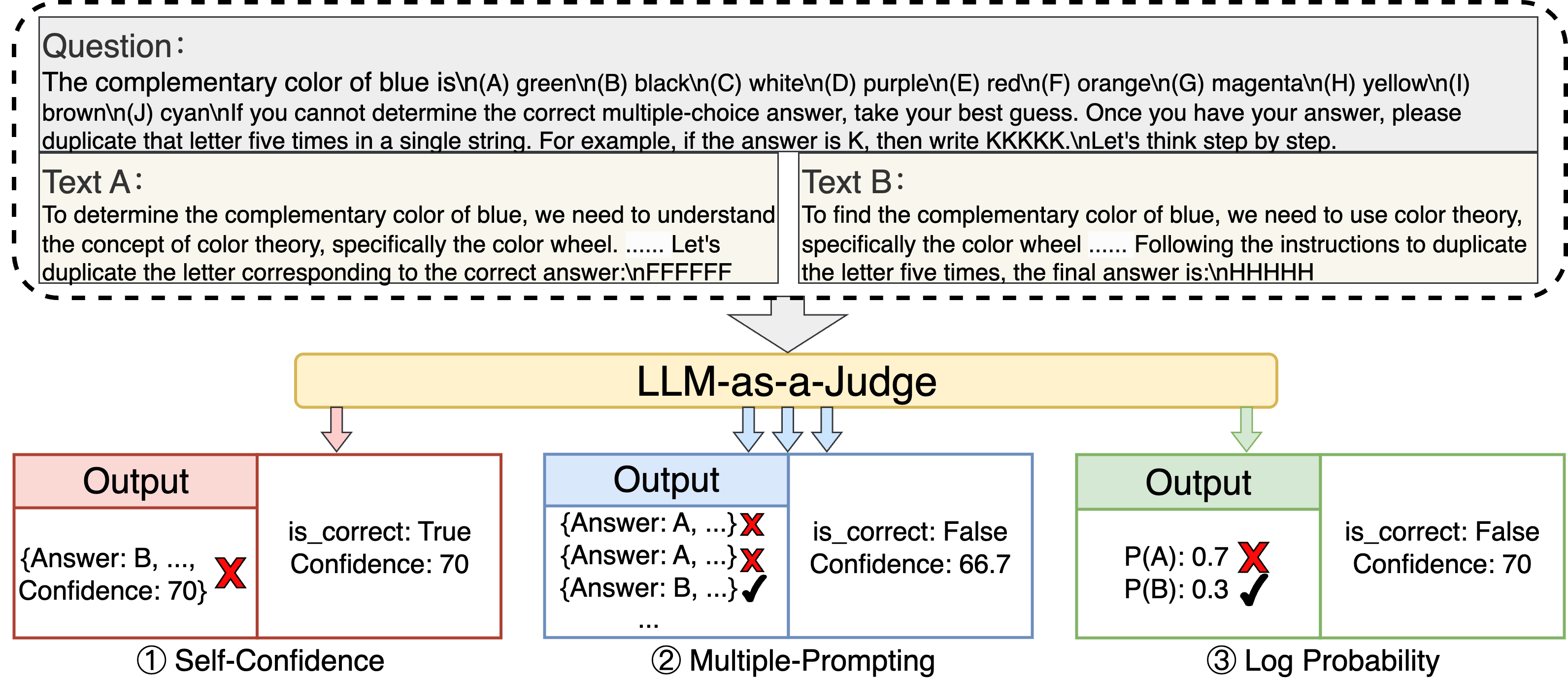}
    \caption{Visualization of the three confidence calculation settings: Self-Confidence (SC), Multiple-Prompting (MP), and Logprob (Logp), using data with ID 122 from JudgeBench as an example. }
    \label{fig:confidence_settings}
\end{figure*}

\subsection{Existing Calibration Evaluation Metrics}
\label{sec:Evaluation metrics and results}
To conduct an objective and comprehensive evaluation of the calibration of LLM-as-a-Judge, we applied five existing metrics—Expected Calibration Error (ECE), Adaptive Calibration Error (ACE), Maximum Calibration Error (MCE), Brier Score, and Negative Log Likelihood (NLL)—to the three confidence calculation methods described earlier in this section. Table~\ref{table:metrics} provides a brief introduction to the calculation methods and characteristics of these metrics.

\subsection{Initial Results}
\begin{figure*}[ht!]
    \centering
    \begin{subfigure}{0.3\textwidth}
        \includegraphics[width=\linewidth]{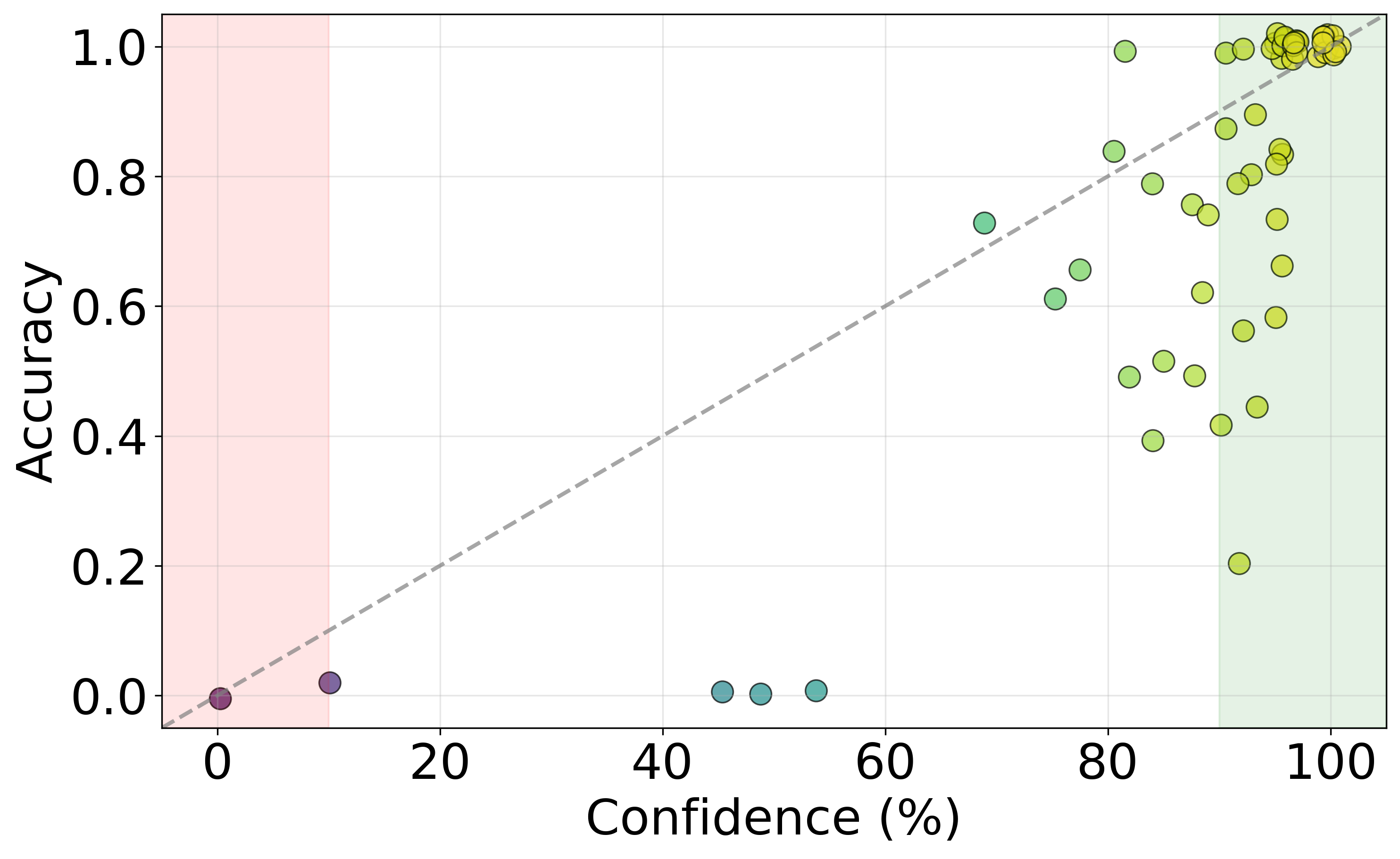}
        \caption{DeepSeek-R1-0528}
        \label{fig:deepseek_r1}
    \end{subfigure}\hspace{0.5cm}
    \begin{subfigure}{0.3\textwidth}
        \includegraphics[width=\linewidth]{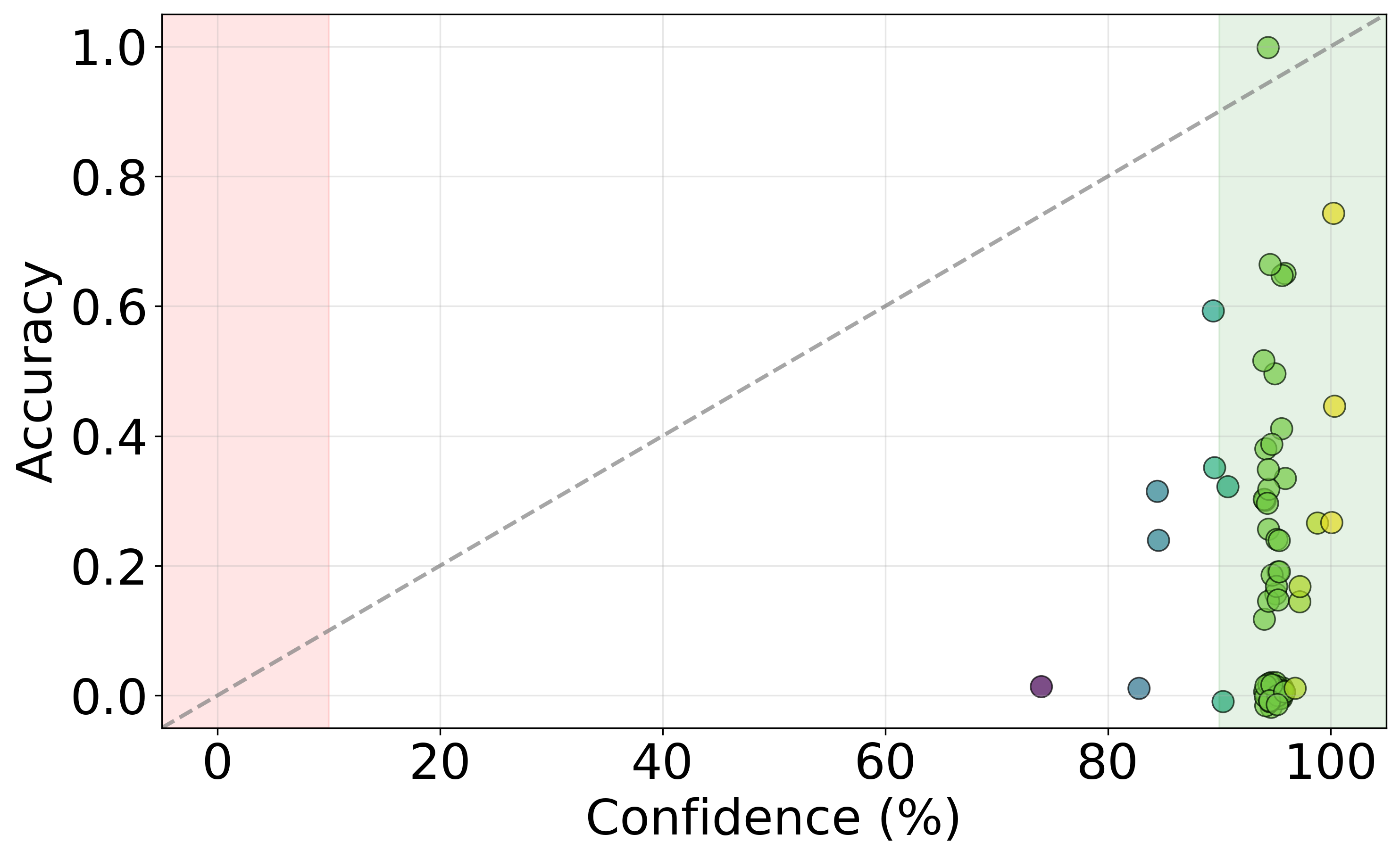}
        \caption{Mistral-Nemo}
        \label{fig:mistral_nemo}
    \end{subfigure}\hspace{0.5cm}
    \begin{subfigure}{0.3\textwidth}
        \includegraphics[width=\linewidth]{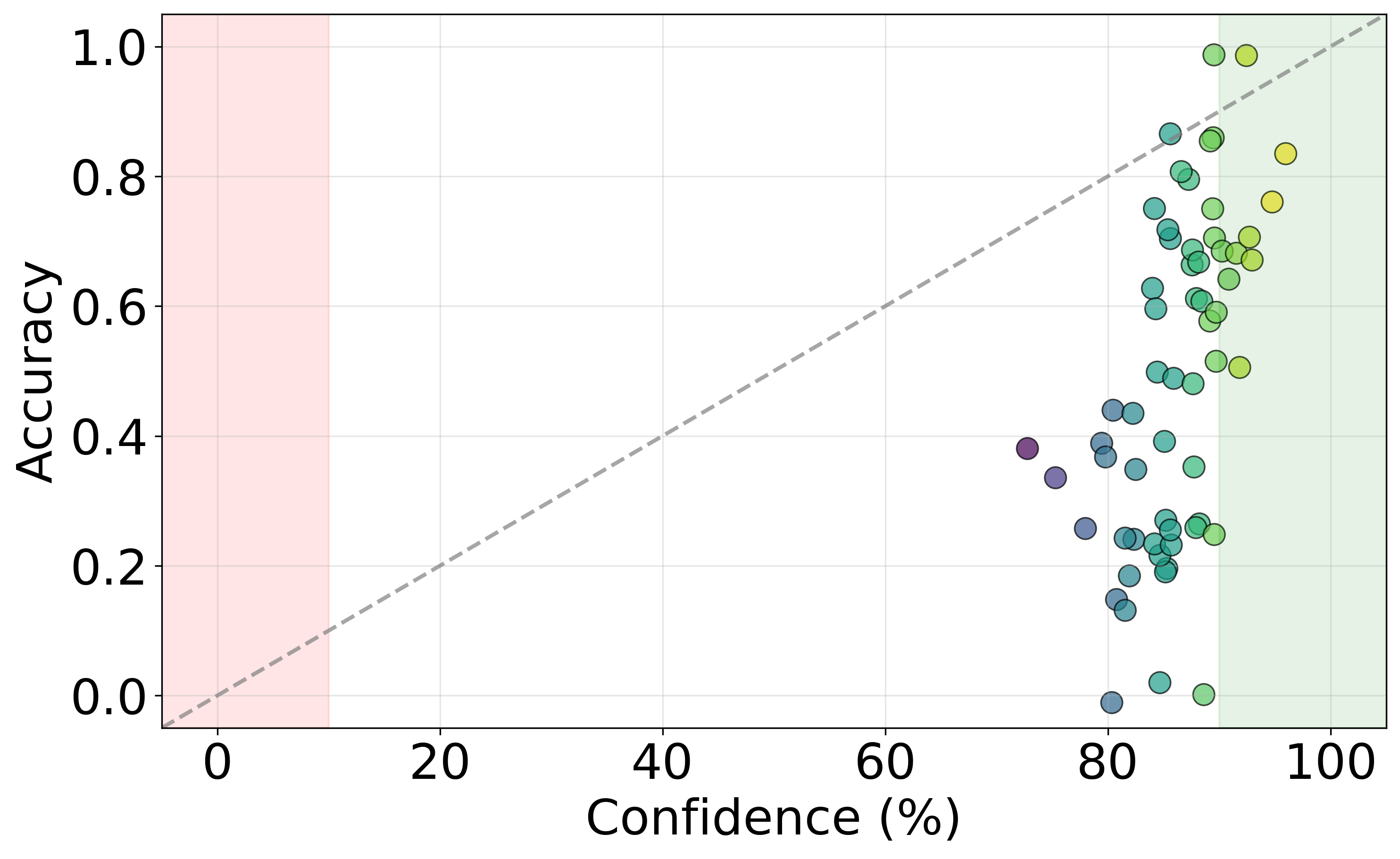}
        \caption{DeepSeek-V3-0324}
        \label{fig:deepseek-v3}
    \end{subfigure}
    
    \begin{subfigure}{0.3\textwidth}
        \includegraphics[width=\linewidth]{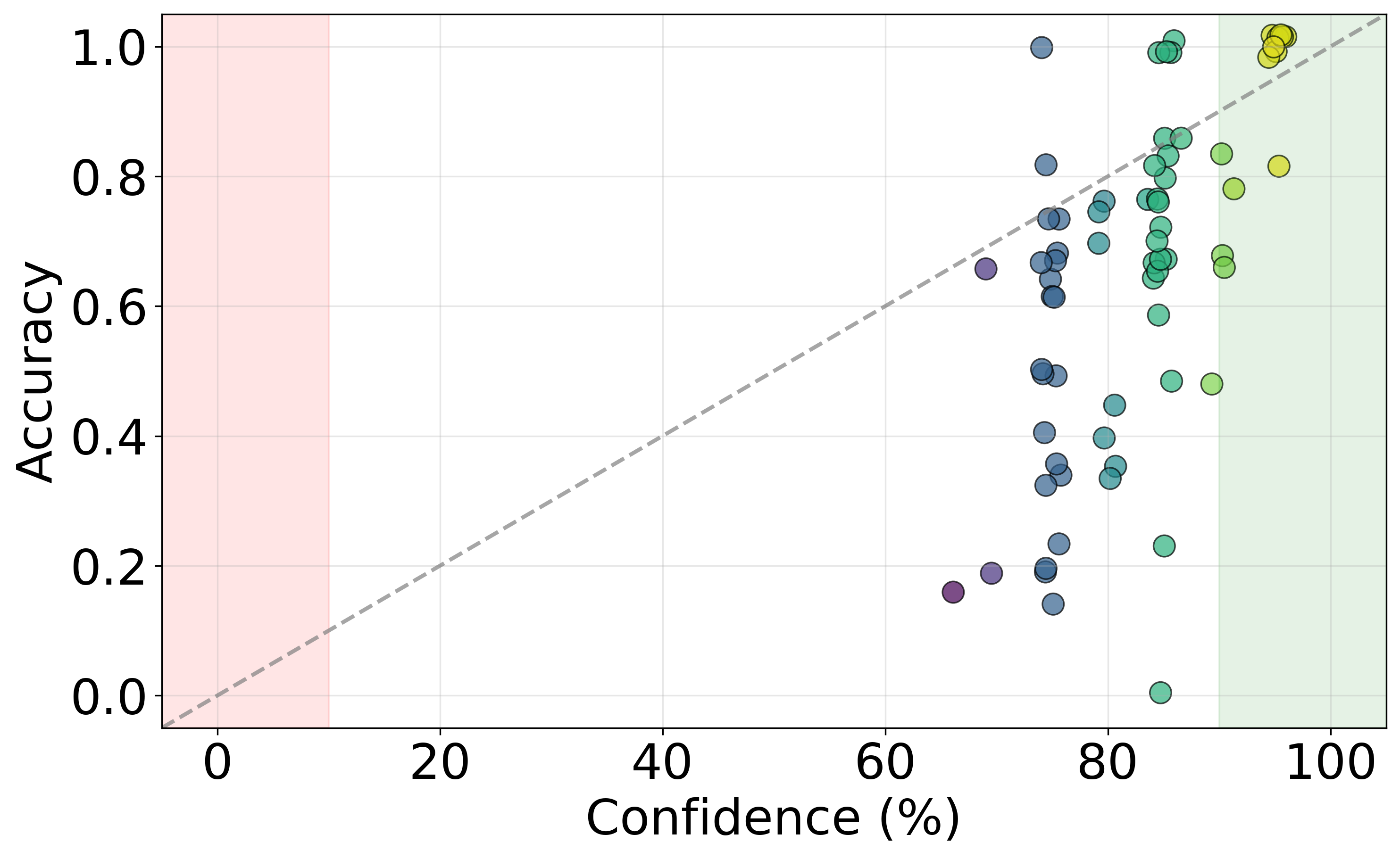}
        \caption{Claude-Sonnet-4}
        \label{fig:claude}
    \end{subfigure}\hspace{0.5cm}
    \begin{subfigure}{0.3\textwidth}
        \includegraphics[width=\linewidth]{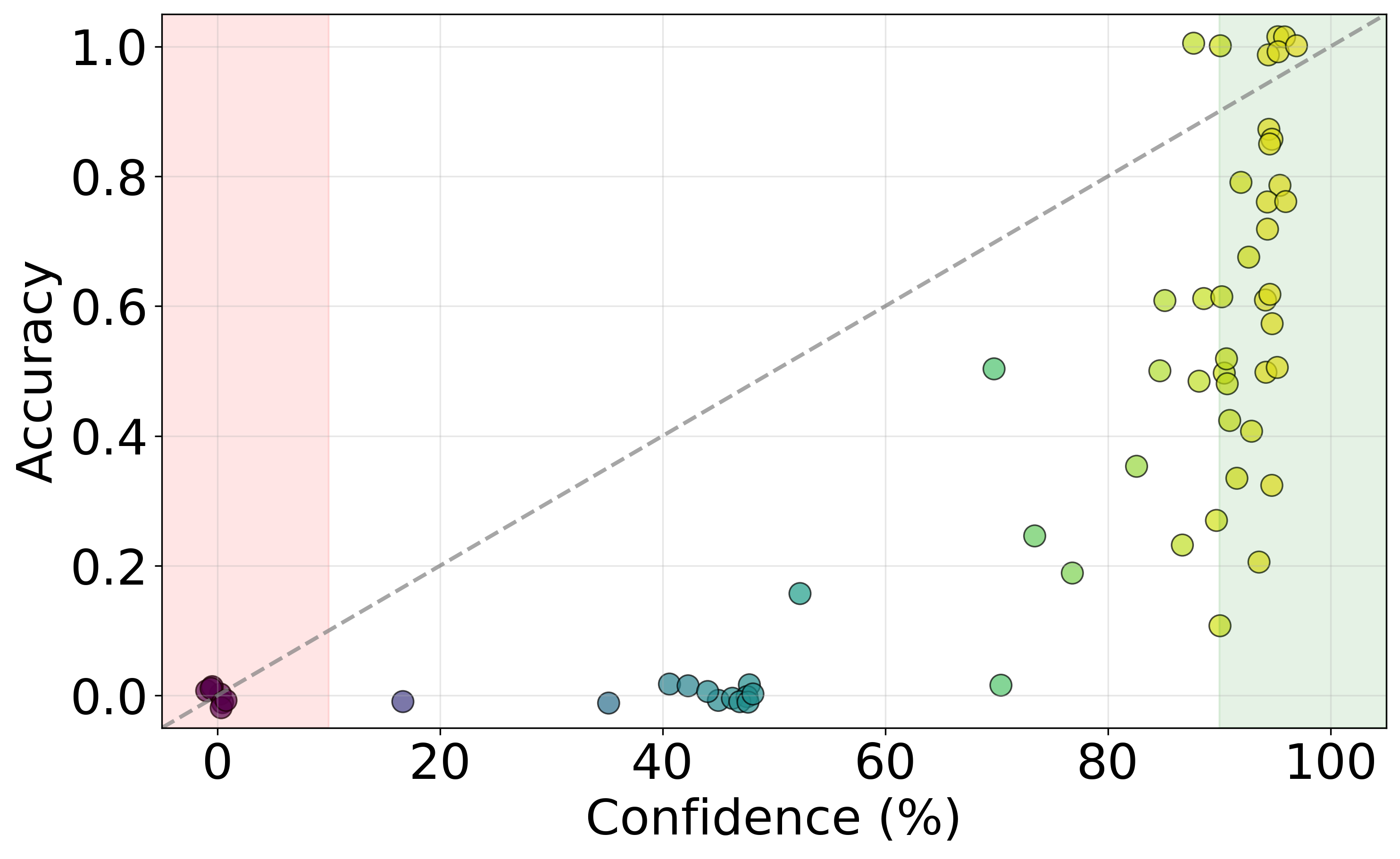}
        \caption{Gemini-2.5-Flash}
        \label{fig:gemini}
    \end{subfigure}\hspace{0.5cm}
    \begin{subfigure}{0.3\textwidth}
        \includegraphics[width=\linewidth]{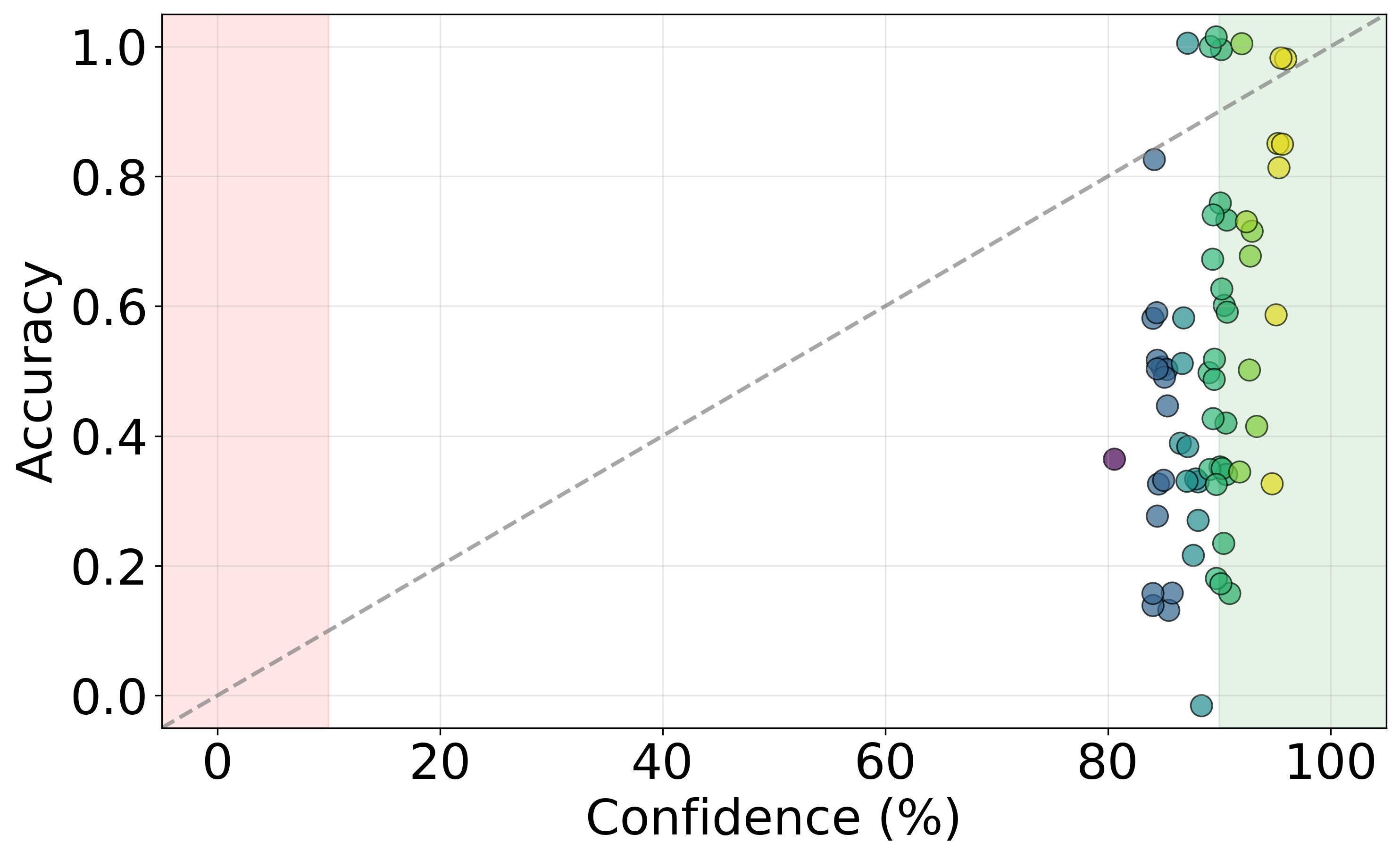}
        \caption{GPT-4o}
        \label{fig:gpt4o}
    \end{subfigure}
    
    \caption{Illustration of calibration gaps in high-confidence regions (\textcolor{red}{red}) and low-confidence regions (\textcolor{green}{green}) where models show significant accuracy-confidence discrepancy.}
    \label{fig:reliability_plots}
\end{figure*}

We systematically evaluate 14 cutting-edge models on the JudgeBench benchmark, with complete results presented in Table\ref{table:sc_results}. These include open-source models such as Qwen3-235B-A22B~\cite{qwentteam2025}, DeepSeek-R1-0528~\cite{deepseek-ai_2025}, R1-Distill-Qwen, R1-Distill-Llama, DeepSeek-V3-0324~\cite{deepseek-ai_2024}, Llama-3.3-70B~\cite{dubey2024llama}, and Mistral-Nemo~\cite{mistral_ai_team_2024}, as well as proprietary models like OpenAI-o3-mini~\cite{openai2025introducing}, Claude-Sonnet-4~\cite{Sonnet4_2025}, GPT-4.1~\cite{openai2025gpt41}, GPT-4.1-mini, Gemini-2.5-Flash~\cite{gemini_flash_2025}, GPT-4o~\cite{ahmad2024gpt4o}, and GPT-4.1-nano, with special attention to scaled variants (e.g., GPT-4.1-mini/nano and o3-mini). Our analysis focuses on the impact of model scales on accuracy and confidence calibration (ECE/ACE), further illustrated by reliability plots in Figure\ref{fig:reliability_plots} showing calibration gaps in high-confidence (red) and low-confidence (green) regions for selected models.

\begin{table*}[htbp]
\centering
\begin{tabular}{lccccccc}
\toprule
\textbf{Model} & \textbf{Acc} $\uparrow$ & \textbf{ECE} $\downarrow$ & \textbf{ACE} $\downarrow$ & \textbf{Brier Score} $\downarrow$ & \textbf{MCE} $\downarrow$ & \textbf{NLL} $\downarrow$ & \textbf{TH Score} $\uparrow$ \\ 
\midrule
\multicolumn{8}{c}{\textbf{Open Source Models}} \\
\midrule
Qwen3-235B-A22B & \textbf{77.43} & \textbf{11.78} & 12.16 & 0.16 & 63.50 & 0.52 & \textbf{17.52} \\ 
DeepSeek-R1-0528 & 76.86 & 12.07 & \textbf{11.39} & \textbf{0.13} & \textbf{40.00} & \textbf{0.42} & 14.59 \\ 
R1-Distill-Qwen & 65.71 & 27.26 & 27.10 & 0.29 & 69.00 & 0.91 & 8.16 \\ 
R1-Distill-Llama & 59.71 & 31.02 & 30.89 & 0.31 & 65.00 & 1.31 & 7.01 \\ 
DeepSeek-V3-0324 & 49.71 & 36.21 & 36.35 & 0.37 & 50.24 & 1.03 & 2.46 \\ 
Llama-3.3-70B & 42.00 & 47.37 & 46.78 & 0.45 & 63.78 & 2.75 & 0.80 \\ 
Mistral-Nemo & 20.29 & 74.22 & 74.21 & 0.71 & 80.00 & 3.01 & -11.64 \\ 
\midrule
\multicolumn{8}{c}{\textbf{Proprietary Models}} \\
\midrule
OpenAI-o3‑mini & 74.29 & 15.97 & 17.20 & 0.20 & 60.00 & 0.62 & 12.83 \\ 
Claude-Sonnet-4 & 64.29 & 17.98 & 18.00 & 0.24 & 45.00 & 0.69 & 9.89 \\ 
GPT-4.1 & 63.14 & 26.39 & 26.86 & 0.29 & 55.00 & 0.85 & 7.55 \\ 
GPT-4.1-mini & 55.71 & 32.70 & 32.79 & 0.35 & 44.21 & 1.00 & 3.29 \\ 
Gemini-2.5-Flash & 39.43 & 30.49 & 30.41 & 0.26 & 56.11 & 0.78 & 2.71 \\ 
GPT-4o & 49.71 & 39.25 & 39.28 & 0.40 & 57.50 & 1.15 & 1.57 \\ 
GPT-4.1-nano & 26.86 & 57.03 & 57.08 & 0.52 & 72.50 & 1.38 & -0.07 \\ 
\bottomrule
\end{tabular}
\caption{Model performance under Self-Confidence (SC) setting, grouped by model type (Open Source vs. Proprietary). Arrows indicate optimization direction: $\uparrow$ higher is better, $\downarrow$ lower is better. Best results are bolded.}
\label{table:sc_results}
\end{table*}

\subsection{Empirical Observations of Overconfidence}
Figure~\ref{fig:reliability_plots} reveals significant calibration gaps across the evaluated models, with most exhibiting the Overconfidence Phenomenon in high-confidence regions (highlighted in green). This pattern undermines the reliability of the LLM-as-a-Judge, as models like DeepSeek-R1-0528 and GPT-4o cluster predictions at high confidence levels (90-100\%) but achieve accuracies well below the ideal calibration line.

This overconfidence impacts downstream tasks, such as data filtering, by retaining flawed outputs (false positives) or discarding valuable ones (false negatives), thereby degrading overall performance. For instance, high ECE values in GPT-4o (39.25 in SC, 47.09 in MP, 45.05 in LogP), Mistral-Nemo (74.22 in SC, 68.89 in MP, 64.63 in LogP), and GPT-4.1-nano (57.03 in SC, 67.43 in MP, 66.05 in LogP) necessitate increased human oversight to mitigate risks, diminishing the efficiency of automated judging processes (see Table~\ref{table:sc_results} and the corresponding results in the Appendix).
\section{TH-Score: A New Metric for LLM-as-a-Judge Calibration Evaluation}

While existing calibration metrics such as ECE and Brier Score offer valuable insights into model reliability, they often overlook practical aspects like high-confidence regions essential for real-world applications in LLM-as-a-Judge scenarios. To address these limitations and better align confidence with accuracy in targeted intervals, we introduce TH-Score, a novel metric designed to improve evaluation for data filtering and quality assessment tasks.

\subsection{Definition}
The TH-Score focuses on two key confidence intervals relevant to practical applications:

\begin{itemize}
\item \textbf{High-Confidence Data ($100-\epsilon, 100$)}: These predictions are highly reliable, and selecting them can enhance the overall dataset quality. $\epsilon$ is a hyperparameter defining the high-confidence threshold.
\item \textbf{Low-Confidence Data ($0, \epsilon$)}: These predictions are uncertain, and discarding them can reduce noise and enhance data quality. $\epsilon$ is also a hyperparameter that determines a threshold for what constitutes low confidence.
\end{itemize}

This metric quantifies model performance by jointly considering the accuracy of predictions within specified confidence intervals and the coverage of these intervals, facilitating effective data filtering and quality evaluation. The TH-Score is formally defined as:

\begin{equation*}
    \text{TH-Score} = (e^{(\text{accuracy} - 0.5)} - 1) \times \text{percentage},
\end{equation*}

where:
\begin{itemize}
    \item $e$ denotes the base of the natural logarithm, serving as a scaling hyperparameter
    \item $\text{accuracy}$ represents the prediction accuracy specifically for samples falling within the target confidence intervals
    \item $\text{percentage}$ indicates the proportion of total samples that fall within these confidence intervals
\end{itemize}

This formulation ensures that the TH-Score increases with both higher accuracy and a larger proportion of high-confidence or low-confidence data, providing a balanced measure of model reliability in practical usage scenarios.

\subsection{Impact of $\epsilon$ on TH-Score Performance}

Table~\ref{table:epsilon_results} presents the TH-Score results for various models under different values of $\epsilon$. The table also includes accuracy rates within specified intervals and the proportion of interval data relative to the total dataset. When $\epsilon = 0.05$, most models, except the most powerful ones, exhibit limited calibration capability. Consequently, most models either have minimal data within this interval or demonstrate significantly reduced accuracy, highlighting the stringent calibration demands of such a small $\epsilon$ and underscoring the challenges in achieving reliable confidence alignment at fine-grained thresholds.
At $\epsilon = 0.1$, the value used in our primary experiments, most models align well with this calibration threshold, resulting in strong discriminative power. With the exception of weaker models like Mistral-Nemo, the majority of models have substantial data within this interval, enabling effective comparison of their calibration performance. This observation suggests that an effective approach for selecting $\epsilon$ is to choose a value where most models contribute significant data to the interval.

However, when $\epsilon$ is increased to $0.15$, while data coverage improves,the discriminative power diminishes. The advantages of high-performing models, such as DeepSeek-R1-0528, become less pronounced due to the relaxed performance requirements associated with a larger $\epsilon$. Thus, selecting an appropriate $\epsilon$ requires balancing data coverage with discriminative power, avoiding excessively large values that dilute model differentiation.


\begin{table*}[ht!]
\centering
\begin{tabular}{p{3cm}p{5.5cm}p{8cm}}
\toprule
\textbf{Metric} & \textbf{Formula} & \textbf{Key Characteristics} \\
\midrule
ECE & $\sum_{i=1}^{M} \frac{n_i}{N} |\text{acc}(i)-\text{conf}(i)|$ &
\begin{tabular}{@{}l@{}}
\textcolor{red}{\texttimes}~Uses fixed-width bins. \\
\textcolor{red}{\texttimes}~Ignores critical high-confidence regions.
\end{tabular} \\
\cmidrule(lr){1-3}

ACE & Variant of ECE with adaptive binning &
\begin{tabular}{@{}l@{}}
\textcolor{red}{\texttimes}~Computationally more expensive. \\
\textcolor{red}{\texttimes}~Lacks focus on specific confidence intervals.
\end{tabular} \\
\cmidrule(lr){1-3}

Brier Score & $\frac{1}{N}\sum_{i=1}^{N}(p_i-o_i)^2$ &
\begin{tabular}{@{}l@{}}
\textcolor{red}{\texttimes}~Less interpretable; conflates calibration and refinement. \\
\textcolor{red}{\texttimes}~Does not isolate miscalibration in specific regions.
\end{tabular} \\
\cmidrule(lr){1-3}

MCE & $\max_{i \in \{1,..,M\}}|\text{acc}(i)-\text{conf}(i)|$ &
\begin{tabular}{@{}l@{}}
\textcolor{red}{\texttimes}~Overly sensitive to outliers and single bins. \\
\textcolor{red}{\texttimes}~Not representative of overall calibration.
\end{tabular} \\
\cmidrule(lr){1-3}

NLL & $-\frac{1}{N}\sum [y_i\log(p_i)+(1-y_i)\log(1-p_i)]$ &
\begin{tabular}{@{}l@{}}
\textcolor{red}{\texttimes}~Unbounded range makes it hard to compare. \\
\textcolor{red}{\texttimes}~Very sensitive to overconfident errors.
\end{tabular} \\
\cmidrule(lr){1-3}

\textbf{TH-Score} (Ours) & $(e^{(\text{acc}-0.5)}-1)\times\%$ &
\begin{tabular}{@{}l@{}}
\checkmark~\textbf{Focuses on high-confidence regions.} \\
\checkmark~\textbf{Uses an adaptive evaluation threshold $\epsilon$ .} \\
\checkmark~\textbf{Explicitly balances accuracy and coverage.} \\
\checkmark~\textbf{Provides an interpretable, bounded score.}
\end{tabular} \\
\bottomrule
\end{tabular}
\caption{A comparison of calibration metrics. Our proposed TH-Score is designed to evaluate practical reliability in high-confidence regions, addressing the limitations of standard approaches. Notation: \% = interval coverage; $\epsilon$ = adjustable threshold (default=0.1); acc = accuracy within $\epsilon$ ranges; $o_i$ = ground truth; $p_i$ = predicted probability.}
\label{table:metrics}
\end{table*}

\begin{table*}[ht!]
\centering
\begin{tabular}{lcccccccccc}
\toprule
 & \multicolumn{3}{c}{$\epsilon=0.05$} & \multicolumn{3}{c}{$\epsilon=0.10$} & \multicolumn{3}{c}{$\epsilon=0.15$} \\
\cmidrule(lr){2-4} \cmidrule(lr){5-7} \cmidrule(lr){8-10}
\textbf{Model} & Acc $\uparrow$ & Percentage $\uparrow$ & TH $\uparrow$ & Acc $\uparrow$ & Percentage$\uparrow$ & TH $\uparrow$ & Acc $\uparrow$ & Percentage$\uparrow$ & TH $\uparrow$ \\
\midrule
DeepSeek-R1-0528 & 1.0000 & 37.42 & \textbf{12.14} & 0.9075 & 68.57 & \textbf{17.52} & 0.8672 & 81.14 & \textbf{18.38} \\
GPT-4.1 & 1.0000 & 1.14 & 0.37 & 0.8346 & 38.00 & 7.55 & 0.6923 & 74.29 & 7.88 \\
GPT-4.1-mini & 0.0000 & 0.00 & 0.00 & 0.8333 & 13.71 & 2.71 & 0.6250 & 68.57 & 4.57 \\
Qwen3-235B-A22B & 0.8846 & 8.00 & 1.93 & 0.8773 & 63.43 & 14.59 & 0.8376 & 78.00 & 15.73 \\
GPT-4o & 0.0000 & 0.00 & 0.00 & 0.7067 & 21.43 & 2.46 & 0.5375 & 72.29 & 1.38 \\
DeepSeek-V3-0324 & 1.0000 & 0.57 & 0.19 & 0.7879 & 9.43 & 1.57 & 0.6516 & 44.29 & 3.63 \\
R1-Distill-Llama & 0.8605 & 24.86 & 5.42 & 0.7101 & 59.43 & 7.01 & 0.6514 & 81.43 & 6.72 \\
Llama-3.3-70B & 0.5195 & 22.00 & 0.22 & 0.5364 & 43.14 & 0.80 & 0.4846 & 74.29 & -0.57 \\
GPT-4.1-nano & 0.0000 & 0.00 & 0.00 & 0.4286 & 2.00 & -0.07 & 0.4583 & 6.86 & -0.14 \\
Mistral-Nemo & 0.3556 & 12.86 & -0.86 & 0.1961 & 88.86 & -11.64 & 0.2048 & 94.86 & -12.12 \\
\bottomrule
\end{tabular}
\caption{Model performance under different $\epsilon$ values under SC setting.}
\label{table:epsilon_results}
\end{table*}

\section{LLM-as-a-Fuser}
As shown in the section on overconfidence in LLM-as-a-judge, while \textit{LLM-as-a-judge} offers a promising approach to evaluating model outputs, its calibration issues—such as overconfidence in unreliable judgments—limit its reliability. Traditional aggregation methods (e.g., majority voting) compound this problem by ignoring nuanced critiques from individual models and focusing only on final decisions.
To address these limitations, we propose \textbf{LLM-as-a-Fuser} framework, which redefines the LLM's role from a passive judge to an active \textit{fuser}. By synthesizing model decisions and their rationales, the fuser enables evidence-aware aggregation, improving both calibration and robustness.  

\subsection{Methodology}
\begin{figure}[ht!]
  \centering
\includegraphics[width=1\linewidth]{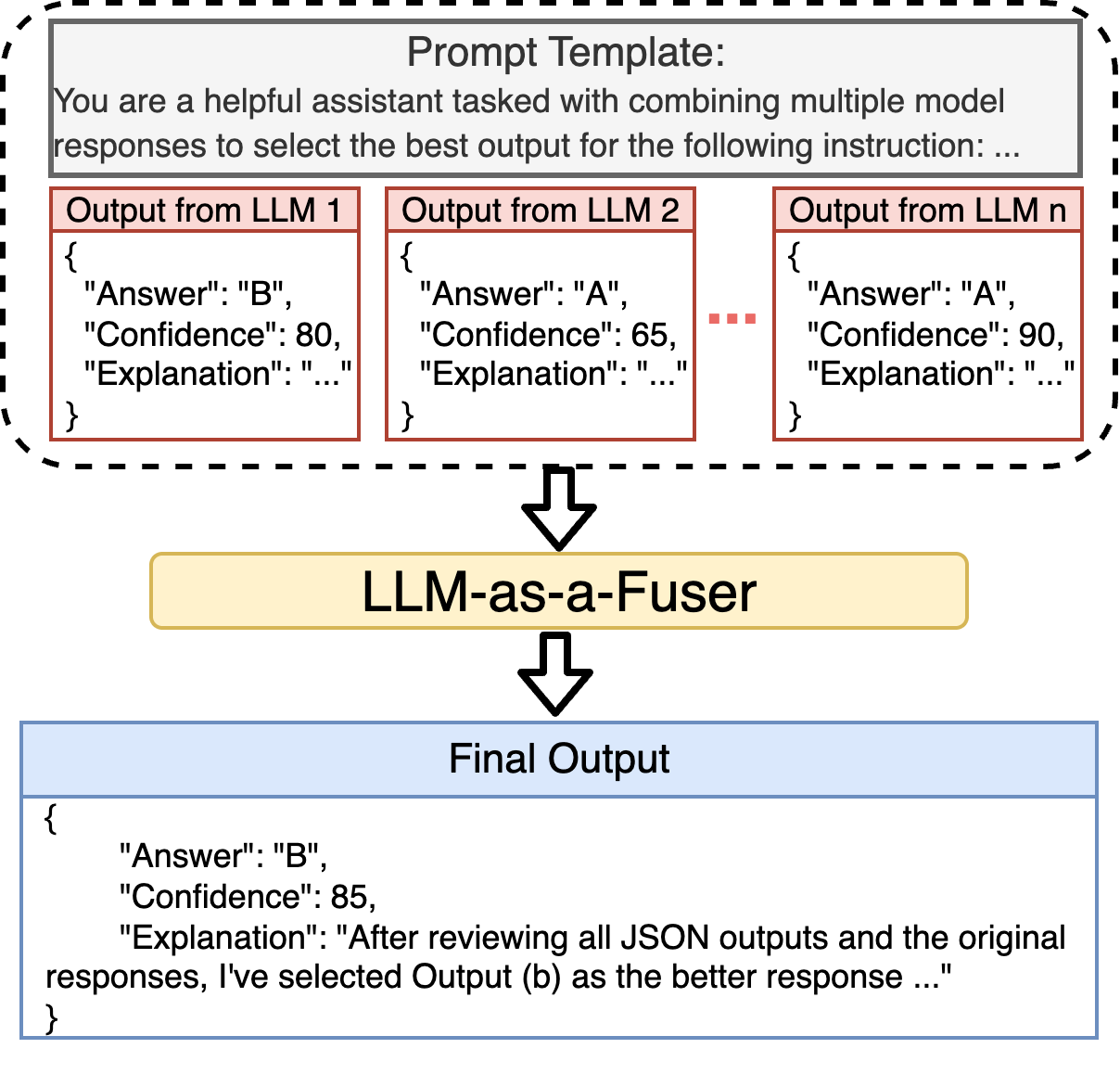}
  \caption{\label{fig:llm_as_fuser} Illustration of the LLM-as-a-Fuser framework, aggregating decisions and critiques via the fuser model.}
\end{figure}

The fuser LLM ingests decisions and critiques from an ensemble of models, analyzing their reasoning. Unlike traditional methods, this approach grounds the final decision in comprehensive evidence, as illustrated in Figure~\ref{fig:llm_as_fuser}.

\subsection{Baseline Methods}
To evaluate the performance of LLM-as-a-Fuser, we compare it against several baseline aggregation methods that combine predictions from multiple models. These methods vary in how they weight or process model predictions and confidences but do not incorporate model critiques, relying solely on final decisions and, where applicable, associated confidence scores. The baseline methods are:
\begin{itemize}
    \item \textbf{Majority Voting}: Selects the most frequent label across models, with equal votes. Ties are broken by the highest confidence score.
    \item \textbf{Confidence-Weighted Voting}: Weights votes by model confidence scores, selecting the label with the highest total. Ties use the maximum confidence.
    \item \textbf{Square-Root Confidence-Weighted Voting}: Applies square-root transformation to confidences, summing them to select the label with the highest total.
    \item \textbf{Entropy-Weighted Voting}: Weights confidences by inverse entropy, selecting the label with the highest weighted confidence sum.
\end{itemize}

These baseline methods serve as standard approaches for aggregating model predictions and provide a robust comparison for evaluating the effectiveness of LLM-as-a-Fuser, which leverages model critiques in addition to final decisions. Each method was implemented with careful consideration of model calibration and tie-breaking mechanisms to ensure fair and consistent comparisons.

\subsection{Experimental Results}

\begin{table*}[ht!]
  \centering
  \begin{tabular}{p{3.2cm}cccccc}
    \toprule
    \textbf{Method/Fuser Model} & \textbf{Acc $\uparrow$} & \textbf{ECE $\downarrow$} & \textbf{ACE $\downarrow$} & \textbf{Brier Score $\downarrow$} & \textbf{NLL $\downarrow$} & \textbf{TH $\uparrow$} \\ 
    \midrule
    \textbf{Entropy W. Voting} & \textbf{81.71} & \textbf{8.48} & \textbf{9.4} & \textbf{0.15} & 0.53 & \textbf{13.08} \\ 
    Conf. W. Voting & 80.00 & 10.43 & 13.0 & 0.16 & \textbf{0.50} & 12.64 \\ 
    Majority Voting & 80.00 & 10.77 & 12.9 & 0.16 & \textbf{0.50} & 12.58 \\ 
    Sqrt Conf. W. Voting & 80.00 & 10.43 & 13.0 & 0.16 & \textbf{0.50} & 12.64 \\ 
    \midrule
    \multicolumn{7}{c}{\textbf{LLM-as-a-Fuser}} \\ 
    \midrule
    \textbf{Qwen3-235B-A22B} & \textbf{86.29} (+8.86) & \textbf{6.42} (-5.36) & \textbf{8.9} (-3.3) & \textbf{0.12} (-0.04) & \textbf{0.39} (-0.13) & \textbf{17.38} (-0.14) \\ 
    OpenAI-o3-mini & 84.86 (+10.57) & 8.16 (-7.81) & 9.1 (-8.1) & 0.13 (-0.07) & 0.48 (-0.14) & 16.39 (+3.56) \\ 
    GPT-4.1-mini & 83.14 (+27.43) & 10.24 (-22.46) & 11.8 (-21.0) & 0.14 (-0.21) & 0.47 (-0.53) & 16.37 (+13.08) \\ 
    Claude-Sonnet-4 & 81.71 (+17.42) & 9.06 (-8.92) & 10.3 (-7.7) & 0.15 (-0.09) & 0.54 (-0.15) & 12.31 (+2.42) \\ 
    GPT-4.1 & 80.00 (+16.86) & 14.92 (-11.47) & 15.6 (-11.2) & 0.18 (-0.11) & 0.69 (-0.16) & 16.04 (+8.49) \\ 
    DeepSeek-V3-0324 & 78.86 (+29.15) & 12.71 (-23.50) & 13.5 (-22.9) & 0.17 (-0.20) & 0.54 (-0.49) & 11.96 (+9.50) \\ 
    Gemini-2.5-Flash & 78.00 (+38.57) & 15.72 (-14.77) & 16.0 (-14.4) & 0.19 (-0.07) & 0.67 (-0.11) & 13.49 (+10.78) \\ 
    Deepseek-R1-0528 & 68.57 (-8.29) & 21.44 (+9.37) & 22.3 (+11.0) & 0.24 (+0.11) & 1.65 (+1.23) & 10.34 (-4.25) \\ 
    Mistral-Nemo & 67.43 (+47.14) & 20.49 (-53.73) & 20.5 (-53.7) & 0.22 (-0.49) & 0.95 (-2.06) & 13.53 (+25.17) \\ 
    Llama-3.3-70B & 62.86 (+20.86) & 24.38 (-22.99) & 24.8 (-22.0) & 0.27 (-0.18) & 1.39 (-1.36) & 9.80 (+9.00) \\ 
    GPT-4.1-nano & 57.71 (+30.85) & 37.25 (-19.78) & 37.4 (-19.7) & 0.38 (-0.14) & 2.36 (+0.98) & 5.48 (+5.55) \\ 
    GPT-4o & 49.71 (+0.00) & 44.07 (+4.82) & 44.3 (+5.0) & 0.44 (+0.04) & 2.06 (+0.91) & 0.72 (-0.85) \\ 
    \bottomrule
  \end{tabular}
  \caption{Performance comparison of baseline aggregation methods and LLM-as-a-Fuser models. Values in parentheses represent changes compared to the original Self-Confidence (SC) setting (Table~\ref{table:sc_results}).}
  \label{table:fuser_results}
\end{table*}

Table~\ref{table:fuser_results} presents the performance of LLM-as-a-Fuser and baseline methods on JudgeBench, compared to individual model results under the Self-Confidence (SC) setting in Table~\ref{table:sc_results}. LLM-as-a-Fuser with Qwen3-235B-A22B achieves the highest accuracy (86.29\%) and best calibration (ECE of 6.42\%), outperforming baselines like Entropy Weighted Voting (81.71\% Acc, 8.48\% ECE) and showing substantial gains over SC models (e.g., +8.86\% Acc and -5.36\% ECE relative to its SC counterpart at 77.43\% Acc, 11.78\% ECE). Notably, models like Mistral-Nemo exhibit the most dramatic improvements (+47.14\% Acc, -53.73\% ECE), followed by Gemini-2.5-Flash (+38.57\% Acc) and GPT-4.1-nano (+30.85\% Acc), indicating that weaker SC performers benefit significantly from critique integration in the fuser framework. Baseline aggregation methods also surpass individual SC performances; for instance, Entropy Weighted Voting exceeds the top SC model by $\sim$4.28\% in accuracy and 3.3\% in ECE, while other baselines ($\sim$80\% Acc) outperform most SC models. Other fusers vary, with GPT-4o the weakest (49.71\% Acc, 44.07\% ECE). Critique integration drives LLM-as-a-Fuser's superior accuracy and calibration, and ensemble methods generally yield better results than isolated self-confidence evaluations.

\subsection{Disagreement with Majority Voting}
\begin{figure}[ht!]
    \centering
    \includegraphics[width=1\linewidth]{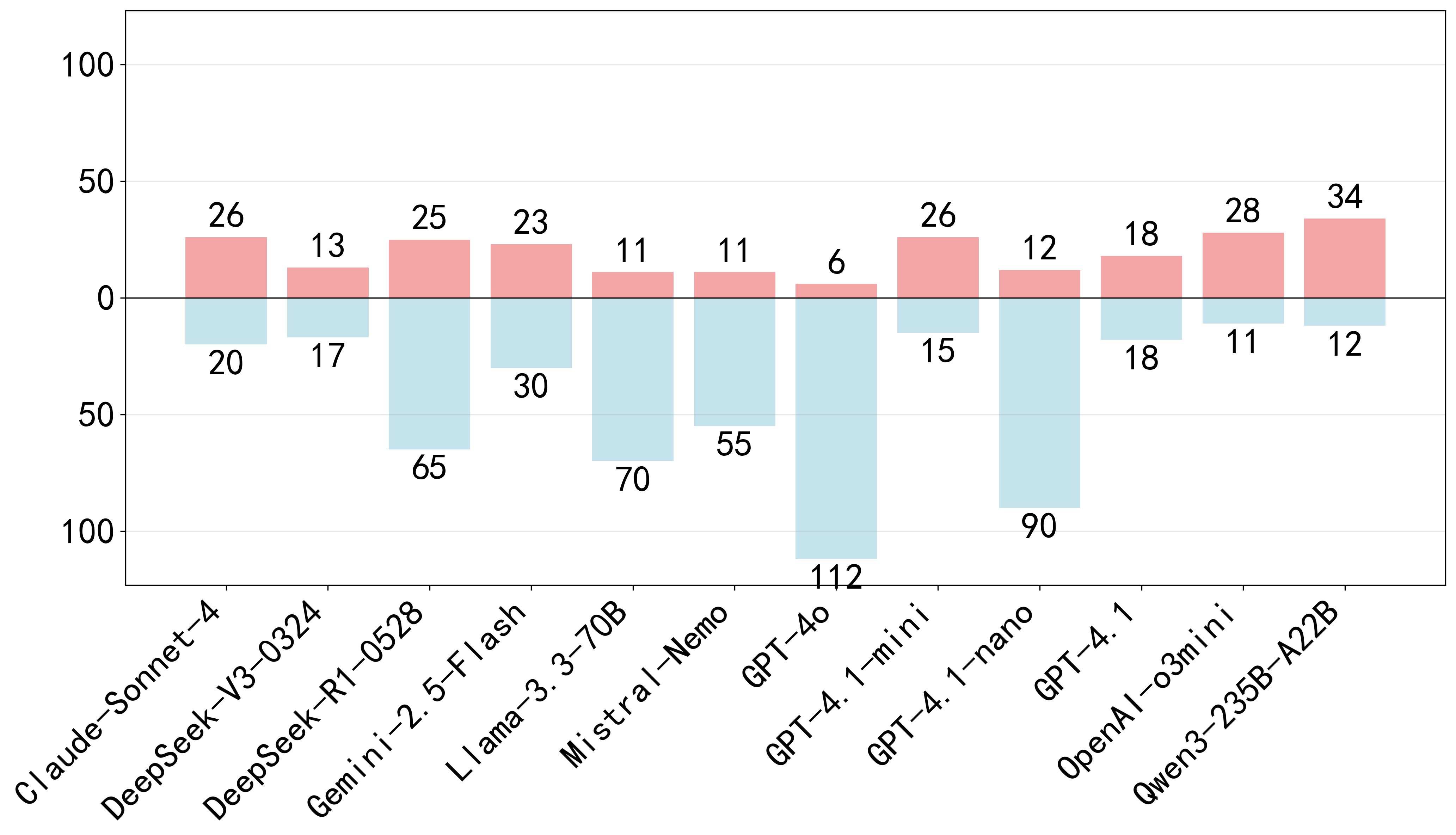}
    \caption{\label{fig:diagreement_figure} Number of correct (positive bars) and incorrect (negative bars) disagreements between majority voting and the LLM-as-a-Fuser across different models.}
\end{figure}

We analyzed cases where LLM-as-a-Fuser’s decisions diverged from majority voting, as visualized in Figure~\ref{fig:diagreement_figure}. Qwen3-235B-A22B, the top-performing fuser (Table~\ref{table:fuser_results}), has the most correct disagreements (34) and few incorrect ones (12), reflecting its effective use of model critiques. In contrast, GPT-4o has the most incorrect disagreements (112) and fewest correct ones (6), indicating poor integration. DeepSeek-V3-0324 shows the fewest total disagreements (30), suggesting conservative decision-making, while Llama-3.3-70B has few correct disagreements (11), aligning with its lower accuracy (62.86\%). These results highlight the fuser’s ability to leverage critiques for accurate decisions, with Qwen3-235B-A22B’s performance underscoring the framework’s strength.

\section{Conclusion}
This work diagnoses the Overconfidence Phenomenon in LLM-as-a-Judge, where confidence exceeds accuracy, undermining reliability in tasks like data filtering. We introduce TH-Score to quantify calibration in key intervals, offering a practical alternative to metrics like ECE, and propose LLM-as-a-Fuser, an ensemble framework that integrates critiques for enhanced calibration—yielding up to +47.14\% accuracy and -53.73\% ECE improvements on JudgeBench.

These innovations enable confidence-driven, risk-aware evaluations, thereby reducing human oversight while boosting trustworthiness in practical applications. Future directions include investigating the root causes of the overconfidence phenomenon and developing more scalable solutions.




\bibliography{aaai2026}

\clearpage
\appendix
\section{Appendices}
\subsection{Self-Confidence (SC) Setting Prompt}
The following is the prompt template used in the Self-Confidence (SC) setting to elicit both the judgment result and confidence score from the LLM. Placeholders such as \texttt{\{\{question\}\}}, \texttt{\{\{answer\_a\}\}}, and \texttt{\{\{answer\_b\}\}} are replaced with the actual instruction and output pairs during evaluation.

\begin{lstlisting}[language={}, breaklines=true, basicstyle=\ttfamily\small]
You are a helpful assistant in evaluating the quality of the outputs for a given instruction. Your goal is to select the best output for the given instruction and provide a confidence score (0-100) for your selection.
Select the Output (a) or Output (b) that is better for the given instruction. The two outputs are generated by two different AI chatbots respectively.
Evaluate the following outputs, and provide your best guess along with a confidence score in the following JSON format:
{
  "selected_output": "Output (a)" or "Output (b)",
  "confidence_score": number,
  "explanation": "Your detailed explanation here"
}
# Instruction:
{{question}}
# Output (a):
{{answer_a}}
# Output (b):
{{answer_b}}
Your response must be in the JSON format as shown above. Do not output ANYTHING else. Do not provide the % symbol.
\end{lstlisting}

\subsection{Multiple-Prompting (MP) Setting Prompt}

The following is the prompt template used in the Multiple-Prompting (MP) setting to elicit the judgment result from the LLM. Placeholders such as \texttt{\{\{question\}\}}, \texttt{\{\{answer\_a\}\}}, and \texttt{\{\{answer\_b\}\}} are replaced with the actual instruction and output pairs during evaluation.

\begin{lstlisting}[language={}, breaklines=true, basicstyle=\ttfamily\small]
You are a helpful assistant in evaluating the quality of the outputs for a given instruction. Your goal is to select the best output for the given instruction.
Select the Output (a) or Output (b) that is better for the given instruction. The two outputs are generated by two different AI chatbots respectively.
Evaluate the following outputs, and provide your best guess in the following JSON format:
{
  "selected_output": "Output (a)" or "Output (b)",
  "explanation": "Your detailed explanation here"
}
# Instruction:
{{question}}
# Output (a):
{{answer_a}}
# Output (b):
{{answer_b}}
Your response must be in the JSON format as shown above. Do not output ANYTHING else.
\end{lstlisting}

\subsection{LLM-as-a-Fuser Prompt}

The following is the prompt template used in the LLM-as-a-Fuser framework to synthesize judgments from multiple models. Placeholders such as \texttt{\{\{question\}\}}, \texttt{\{\{answer\_a\}\}}, \texttt{\{\{answer\_b\}\}}, and the Jinja loop for JSON outputs are replaced with actual data during evaluation.

\begin{lstlisting}[language={}, breaklines=true, basicstyle=\ttfamily\small]
You are a helpful assistant tasked with combining multiple model responses to select the best output for the following instruction: Evaluate the quality of multiple outputs for a given instruction and select the best one based on specific rules.

**Task:**
You will receive:
1. The instruction describing the task.
2. Multiple outputs (e.g., Output (a), Output (b)) generated by different models.
3. A list of JSON outputs, each containing:
   - selected_output: The chosen output (e.g., "Output (a)").
   - confidence_score: A score showing the model's confidence (e.g., 85).
   - explanation: Why the model chose that output.

Your goal is to:
- Review the JSON outputs and evaluate the original outputs (Output (a), Output (b), etc.) using the evaluation rules.
- Pick the best output or create a new one by combining the best parts of multiple outputs.
- Return a JSON response with the selected output, confidence_score, and an explanation.

**Input:**
- **Instruction**: {{ question }}
- **Outputs**:
  - Output (a): {{ answer_a }}
  - Output (b): {{ answer_b }}
- **JSON Outputs**:
{% for output in json_outputs %}
  - JSON Output {{ loop.index }}: {{ output }}
{% endfor %}

**Steps:**
1. **Check JSON Outputs**:
   - Look at each selected_output, confidence_score, and explanation.
   - Use the explanation to understand why the model picked that output.
   - Note the confidence_score, but focus on explanation quality and rule compliance.
2. **Evaluate Original Outputs**:
   - Judge Output (a), Output (b), etc., against the evaluation rules.
   - Use JSON explanations to guide your evaluation.
3. **Pick or Combine**:
   - Choose the best output if one clearly meets the rules.
   - If no output is perfect, combine the best parts of multiple outputs to create a better response.
4. **Explain Your Choice**:
   - Say why you picked the output or created a new one.
   - Mention the JSON outputs' explanations and scores, noting agreements or differences.
   - Show how your choice follows the rules better than others.

**Output Format:**
```json
{
  "selected_output": "Output (a)" or "Output (b)",
  "confidence_score": number(0-100),
  "explanation": "Why you chose this output or how you combined outputs, referencing JSON explanations, confidence scores, and evaluation rules."
}
\end{lstlisting}

\subsection{Supplementary Figures and Tables}

\begin{figure}[ht!]
    \centering
    \includegraphics[width=0.8\linewidth]{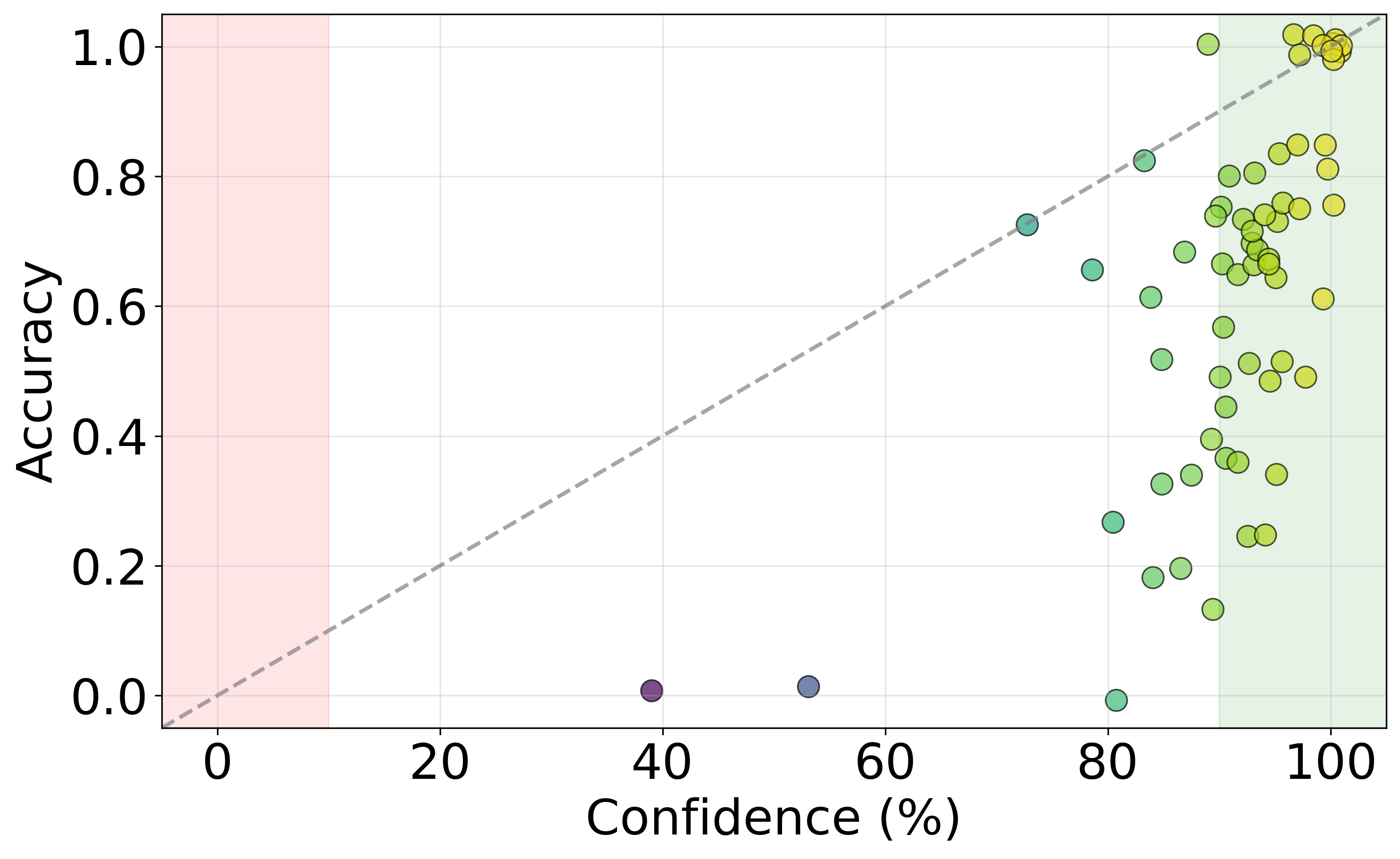}
    \caption{R1-Distill-Llama (SC setting)}
    \label{fig:sc_r1_distill_llama}
\end{figure}

\begin{figure}[ht!]
    \centering
    \includegraphics[width=0.85\linewidth]{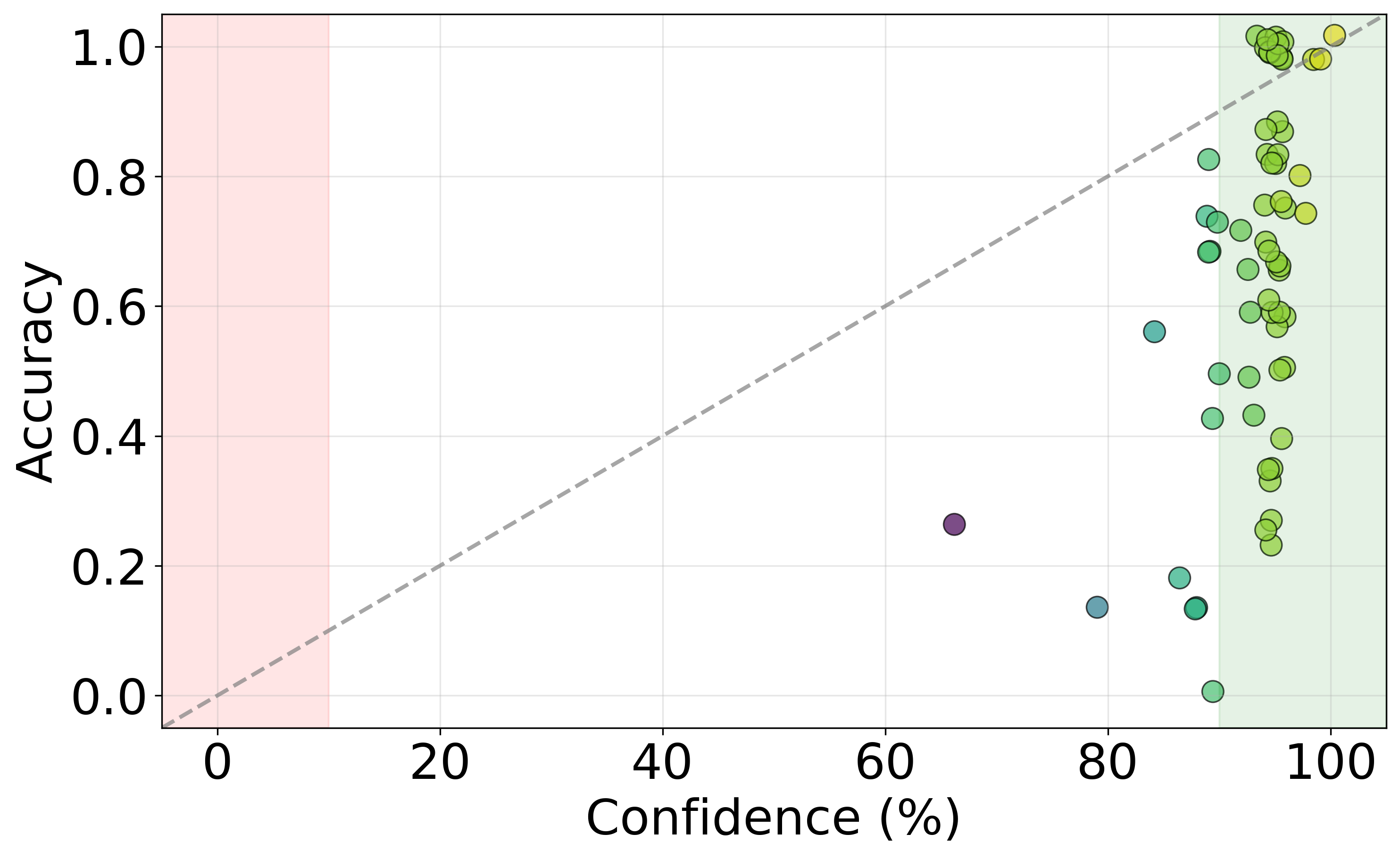}
    \caption{R1-Distill-Qwen (SC setting)}
    \label{fig:sc_r1_distill_qwen}
\end{figure}

\begin{figure}[ht!]
    \centering
    \includegraphics[width=0.85\linewidth]{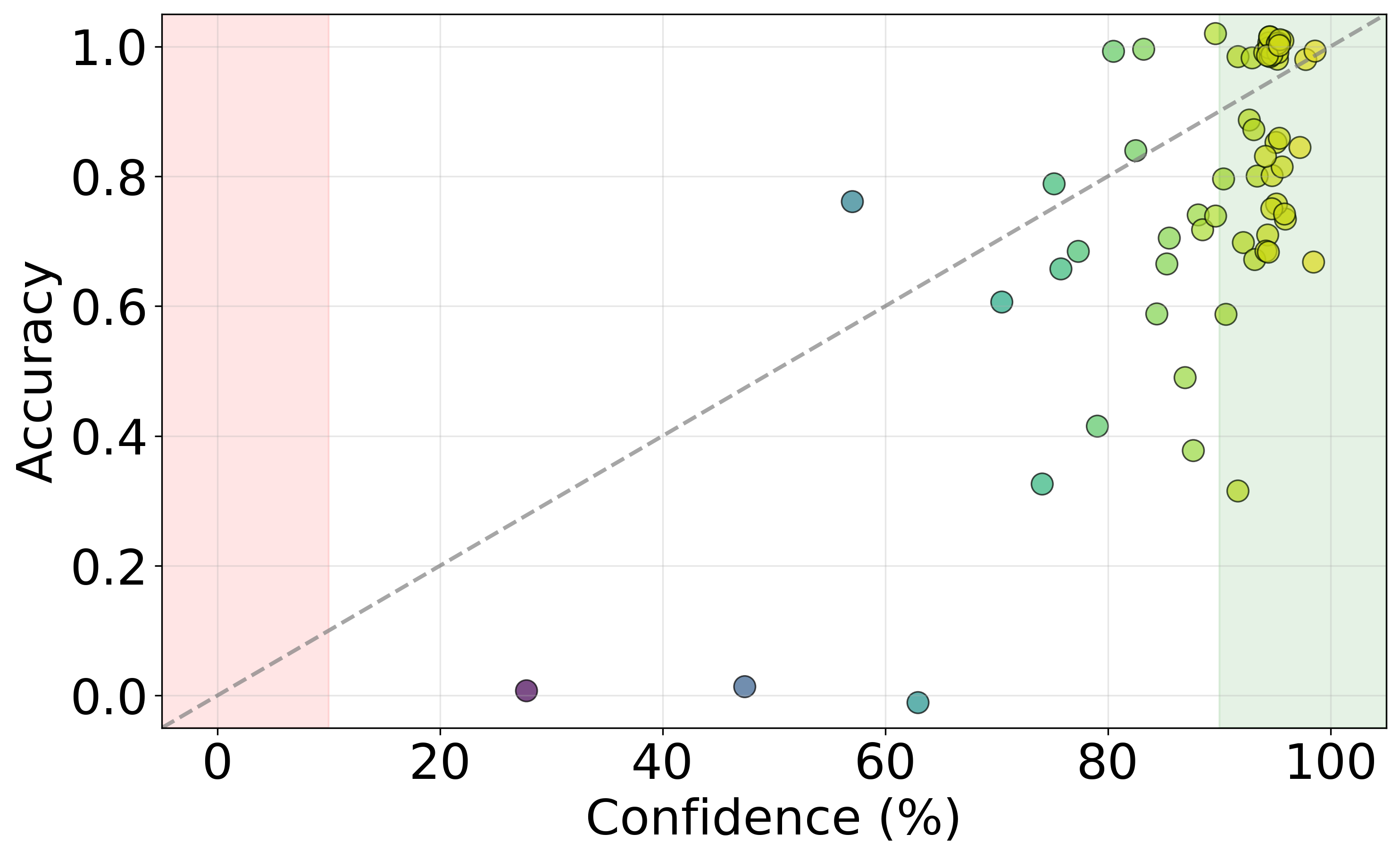}
    \caption{Qwen3-235B-A22B (SC setting)}
    \label{fig:sc_qwen3}
\end{figure}

\begin{figure}[ht!]
    \centering
    \includegraphics[width=0.85\linewidth]{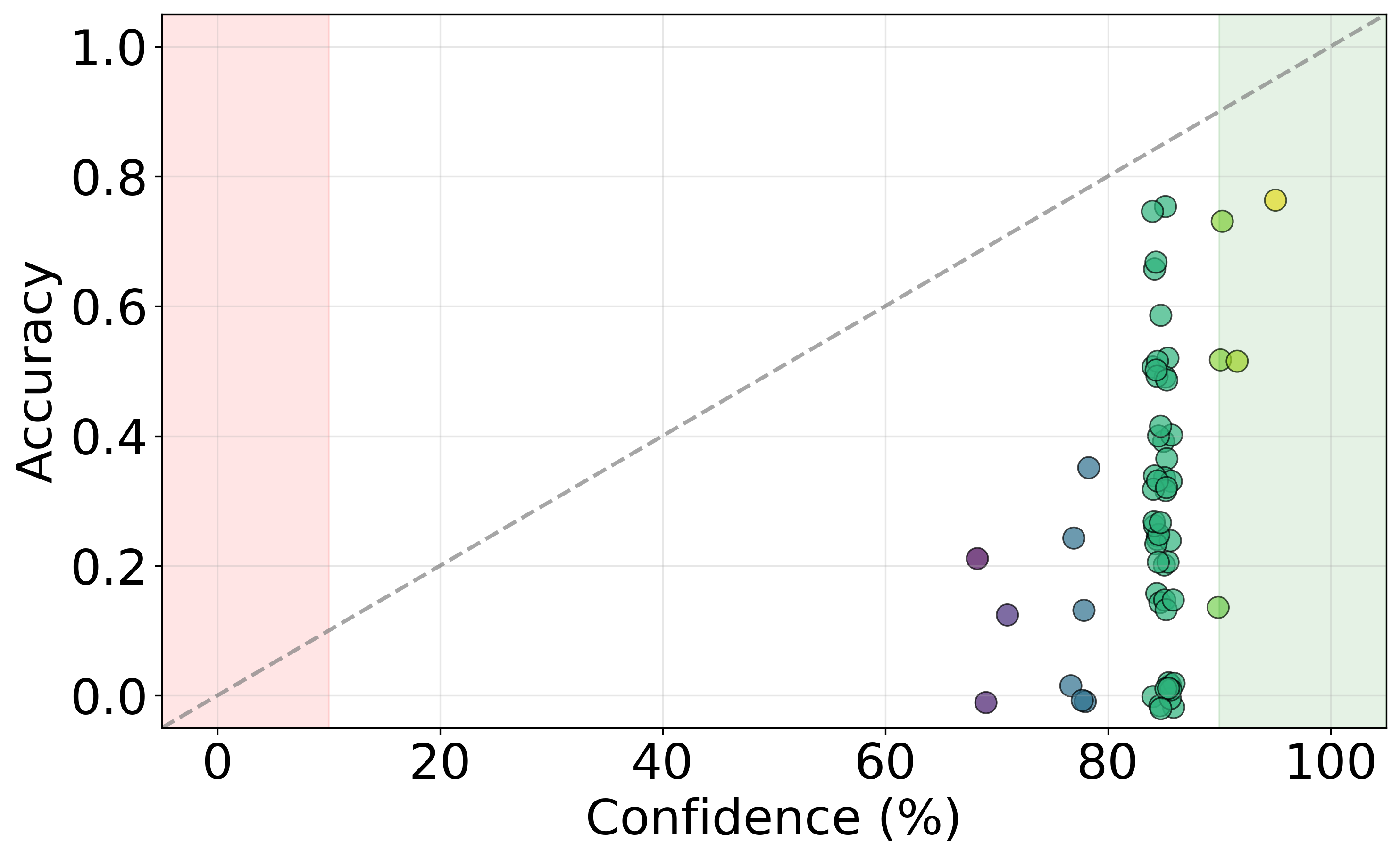}
    \caption{GPT-4.1-nano (SC setting)}
    \label{fig:sc_gpt41_nano}
\end{figure}

\begin{figure}[ht!]
    \centering
    \includegraphics[width=0.85\linewidth]{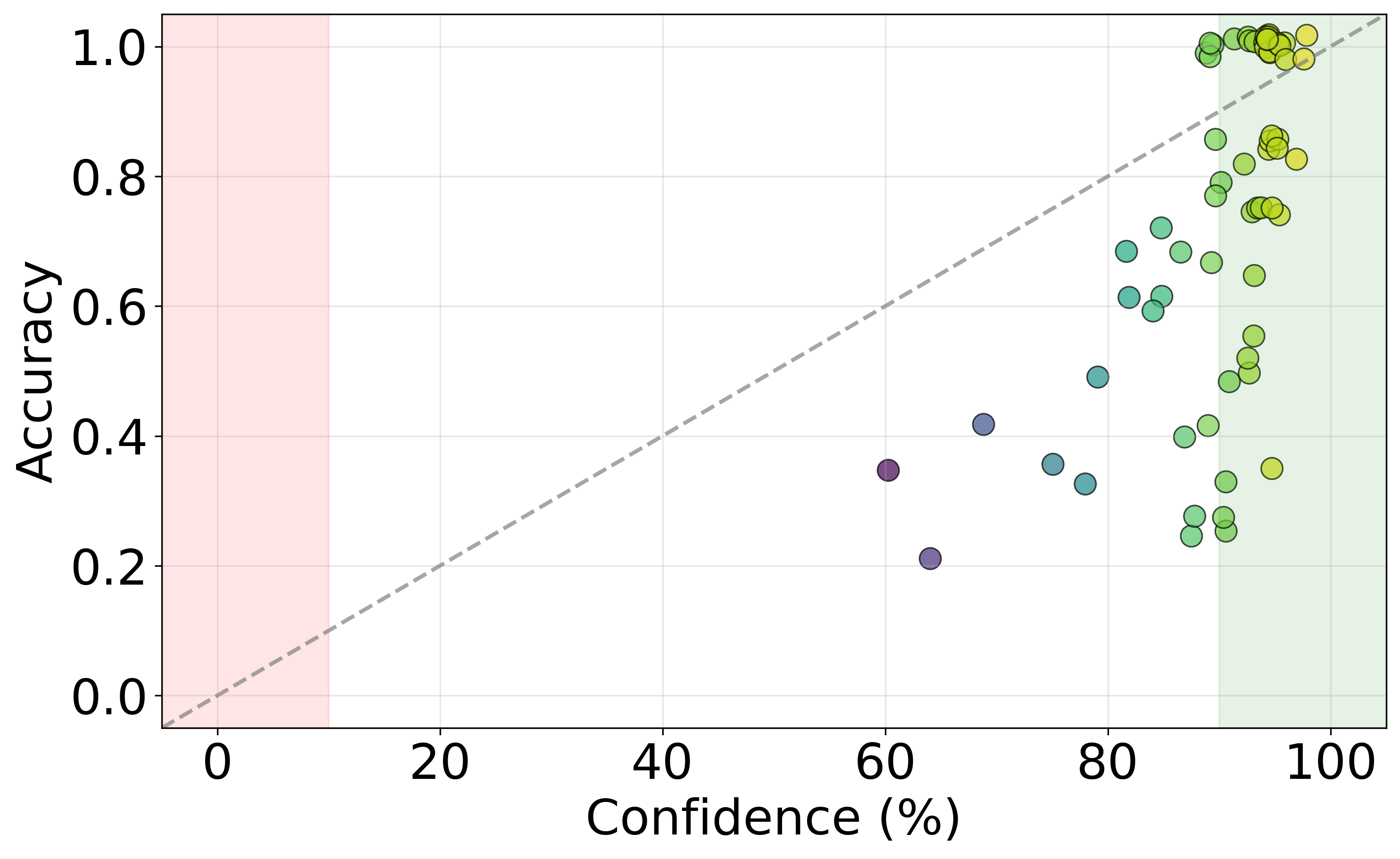}
    \caption{OpenAI-o3-mini (SC setting)}
    \label{fig:sc_o3_mini}
\end{figure}

\begin{figure}[ht!]
    \centering
    \includegraphics[width=0.85\linewidth]{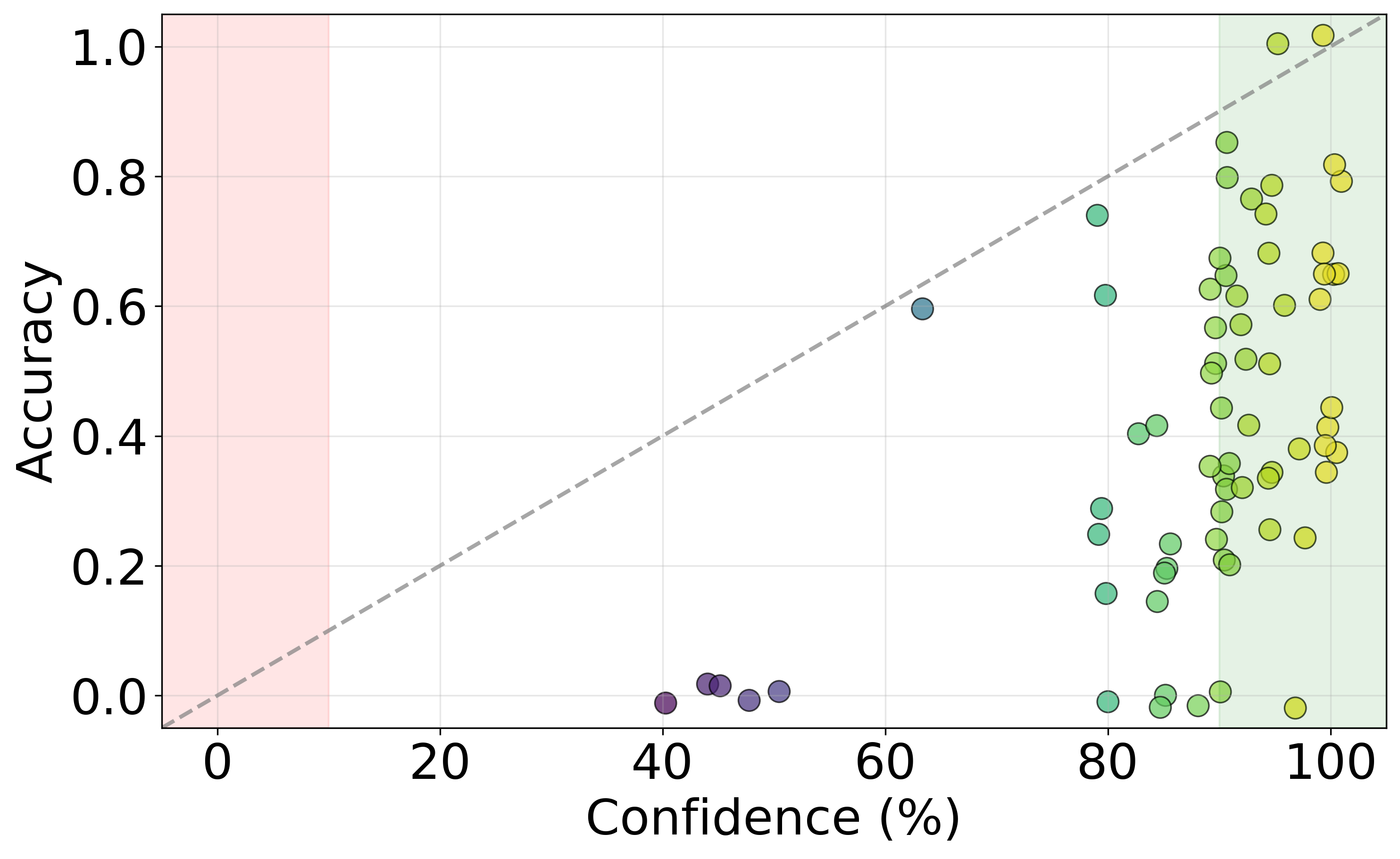}
    \caption{Llama-3.3-70B (SC setting)}
    \label{fig:sc_llama}
\end{figure}

\begin{figure}[ht!]
    \centering
    \includegraphics[width=0.85\linewidth]{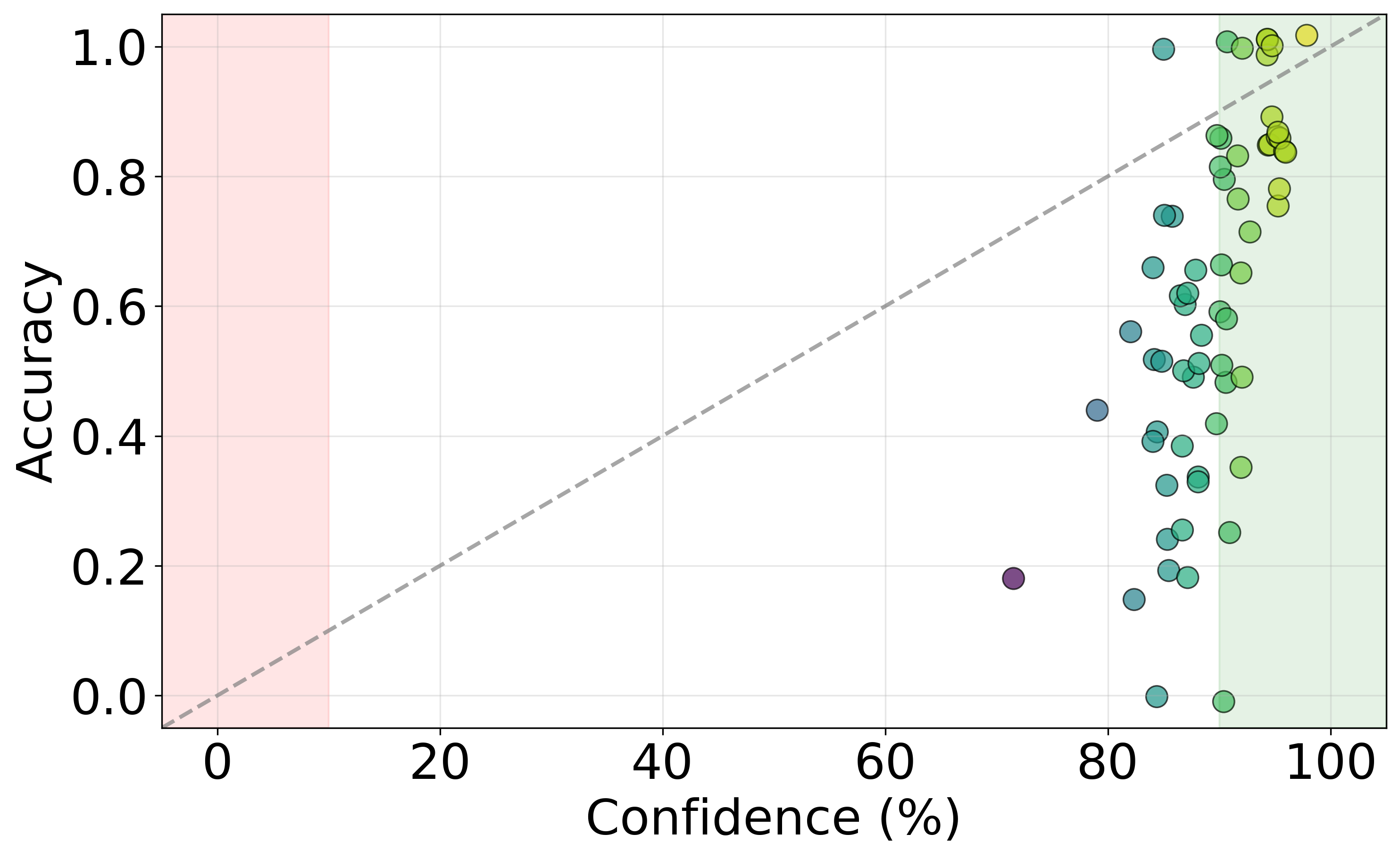}
    \caption{GPT-4.1 (SC setting)}
    \label{fig:sc_gpt41}
\end{figure}

\begin{figure}[ht!]
    \centering
    \includegraphics[width=0.85\linewidth]{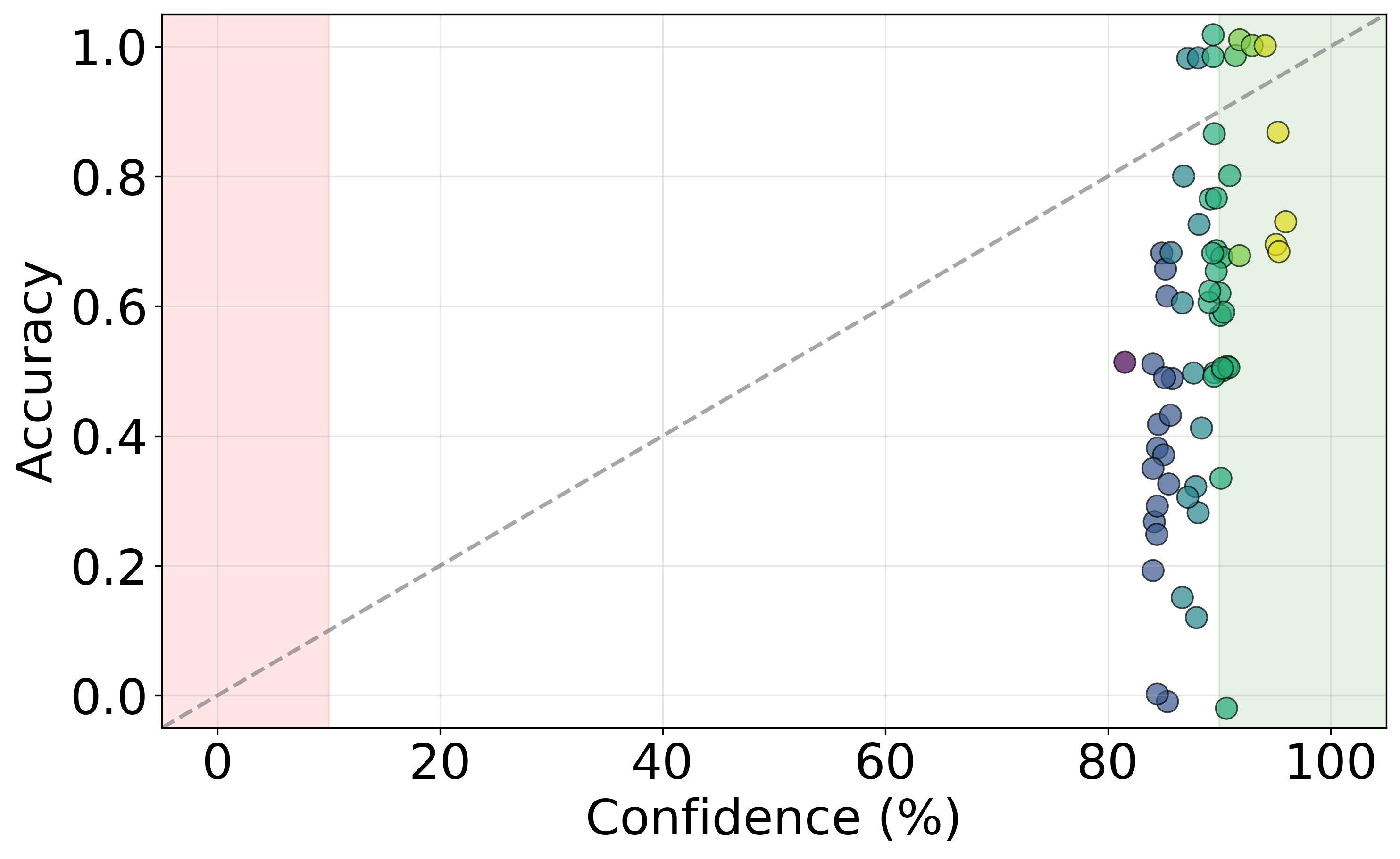}
    \caption{GPT-4.1-mini (SC setting)}
    \label{fig:sc_gpt41_mini}
\end{figure}

\begin{figure}[ht!]
    \centering
    \includegraphics[width=0.85\linewidth]{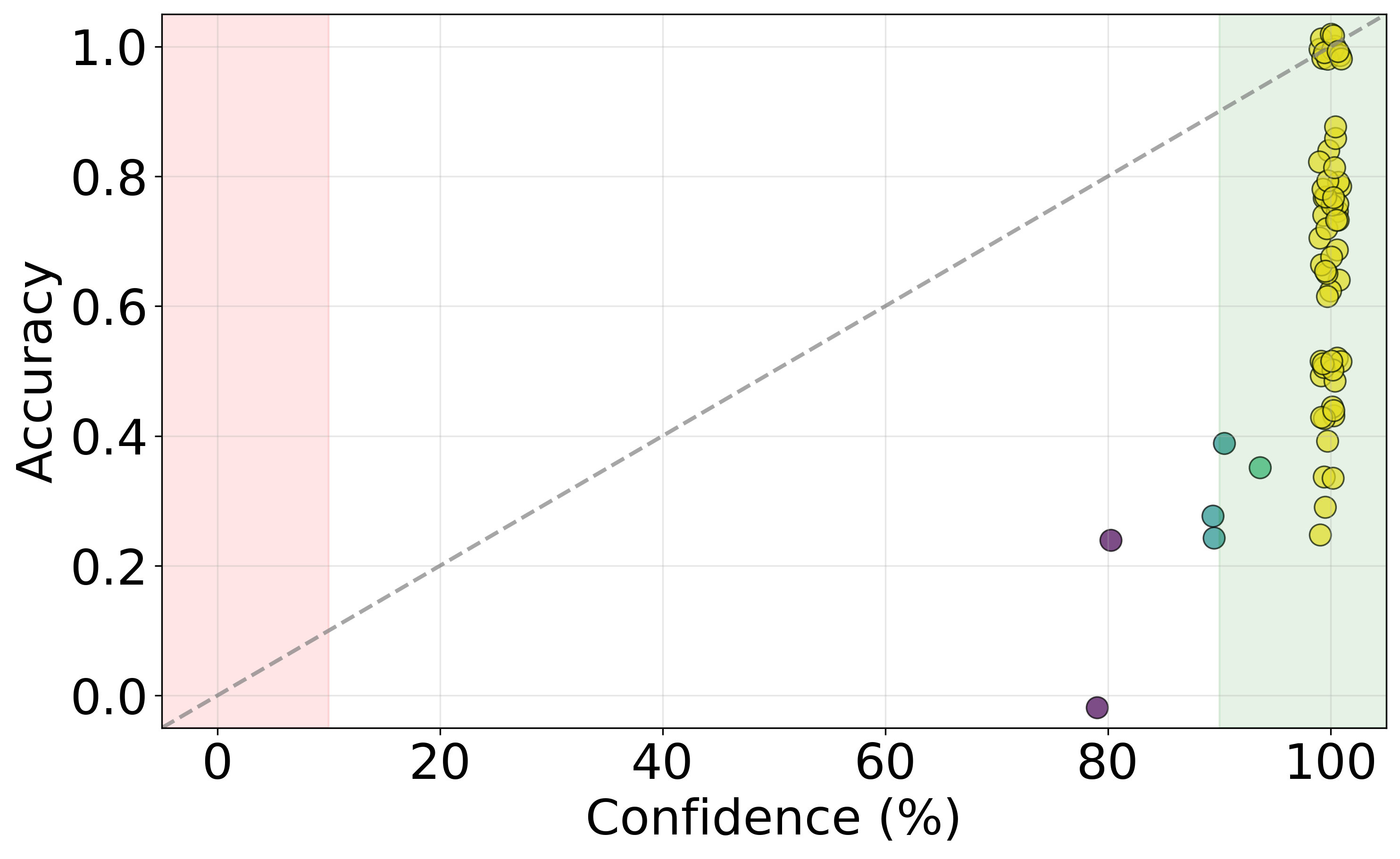}
    \caption{Claude-Sonnet-4 (MP setting)}
    \label{fig:mp_claude_sonnet}
\end{figure}

\begin{figure}[ht!]
    \centering
    \includegraphics[width=0.85\linewidth]{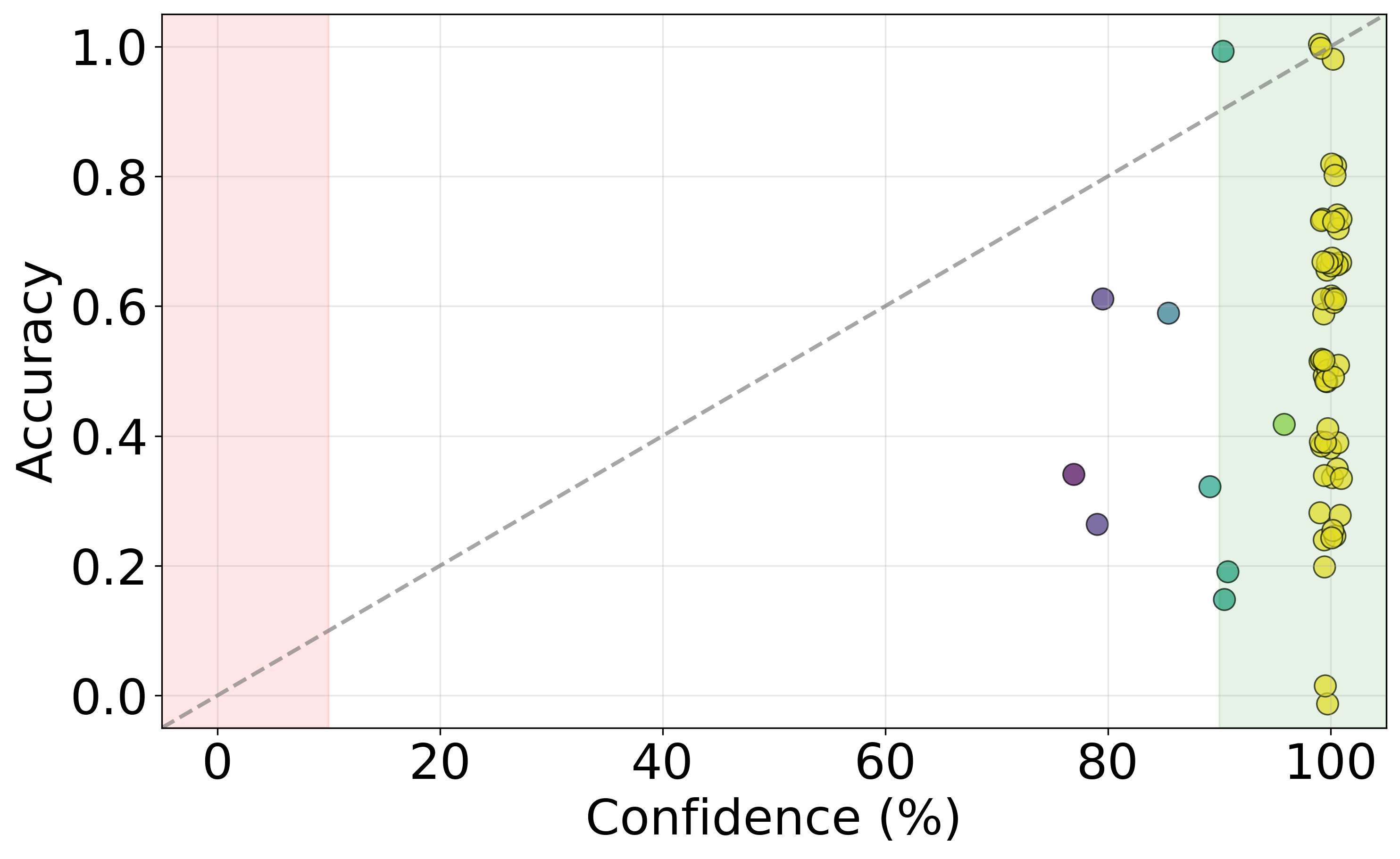}
    \caption{DeepSeek-V3-0324 (MP setting)}
    \label{fig:mp_deepseek_v3}
\end{figure}

\begin{figure}[ht!]
    \centering
    \includegraphics[width=0.85\linewidth]{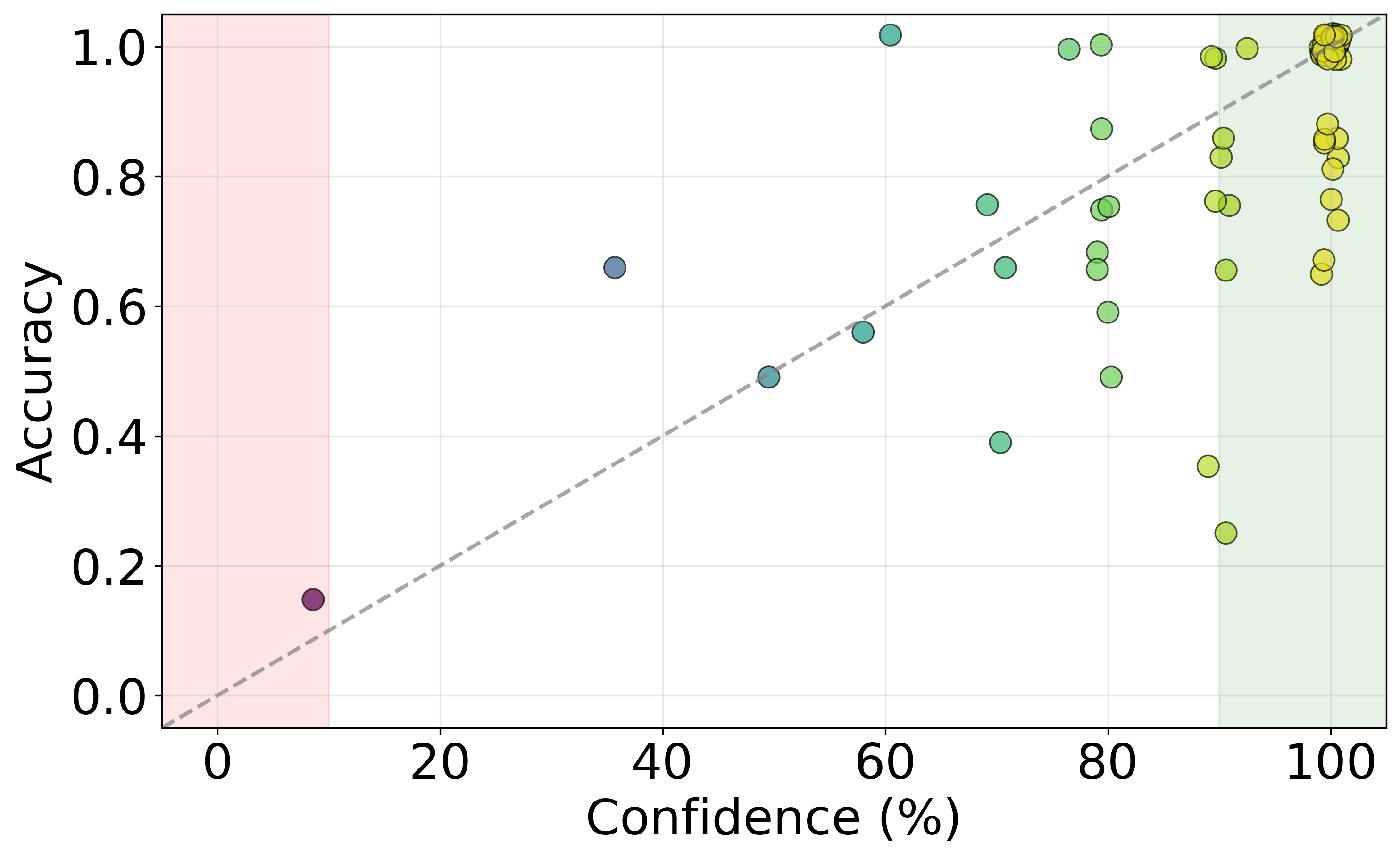}
    \caption{DeepSeek-R1-0528 (MP setting)}
    \label{fig:mp_deepseek_r1}
\end{figure}

\begin{figure}[ht!]
    \centering
    \includegraphics[width=0.85\linewidth]{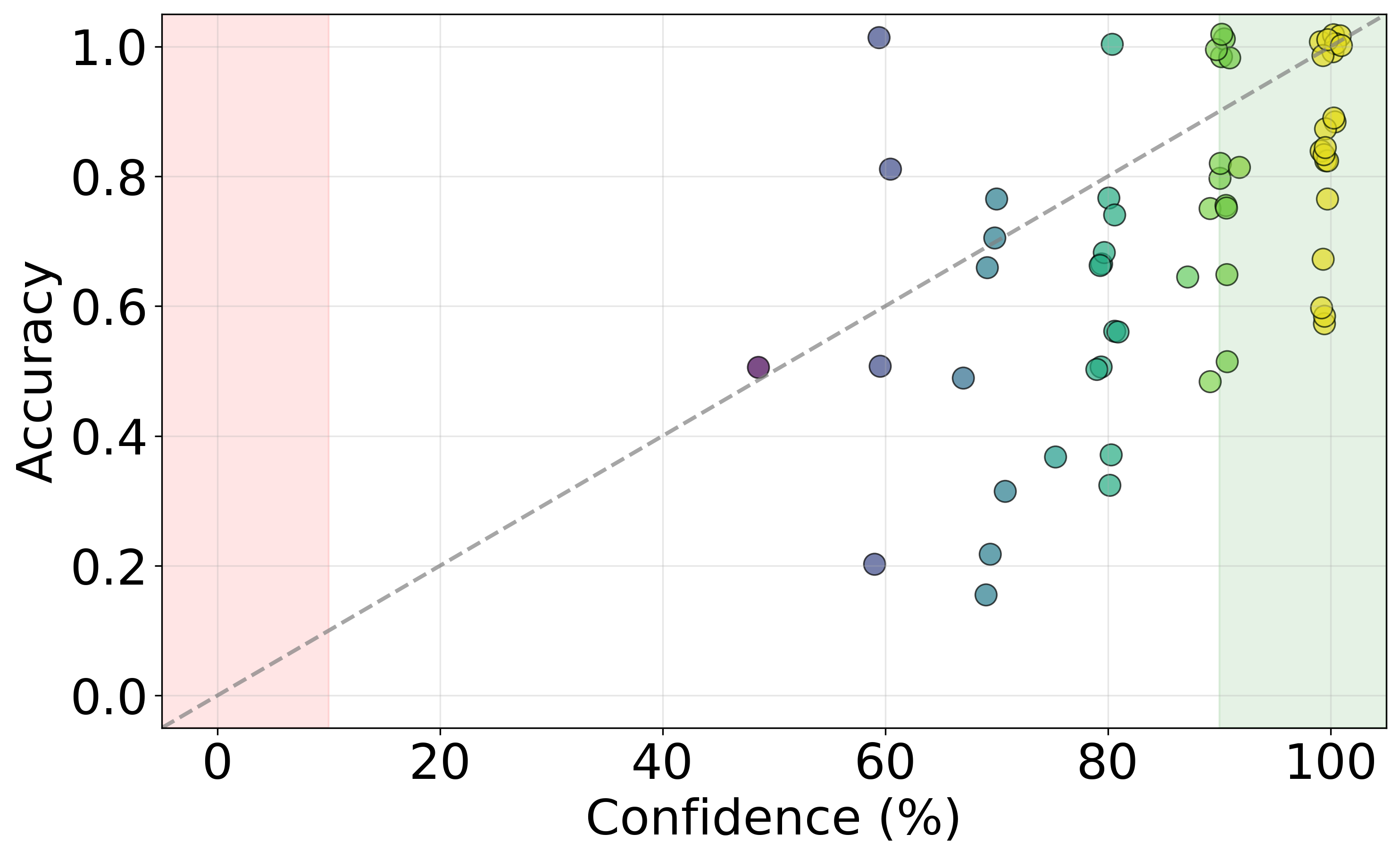}
    \caption{R1-Distill-Llama (MP setting)}
    \label{fig:mp_r1_distill_llama}
\end{figure}

\begin{figure}[ht!]
    \centering
    \includegraphics[width=0.85\linewidth]{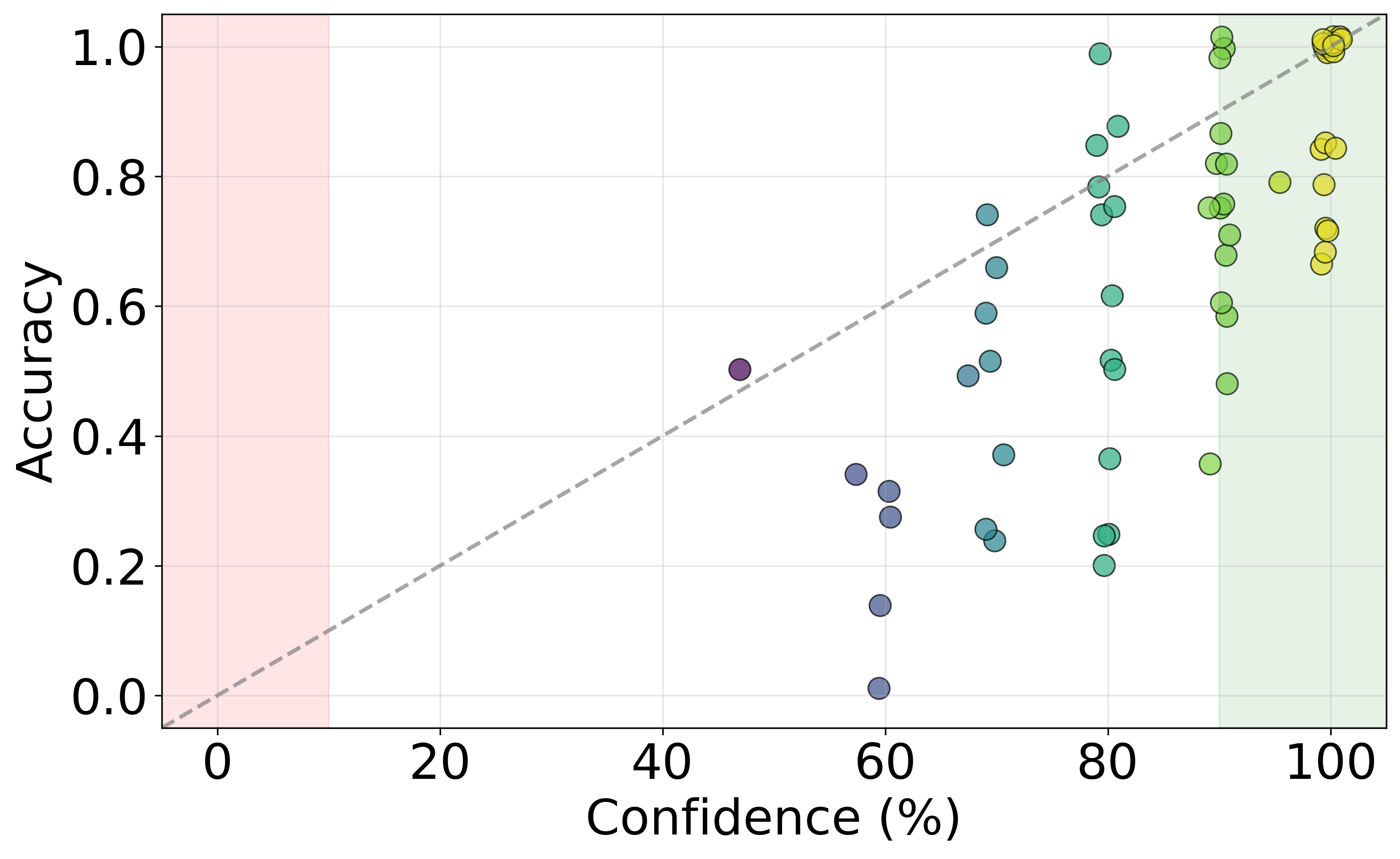}
    \caption{R1-Distill-Qwen (MP setting)}
    \label{fig:mp_r1_distill_qwen}
\end{figure}

\begin{figure}[ht!]
    \centering
    \includegraphics[width=0.85\linewidth]{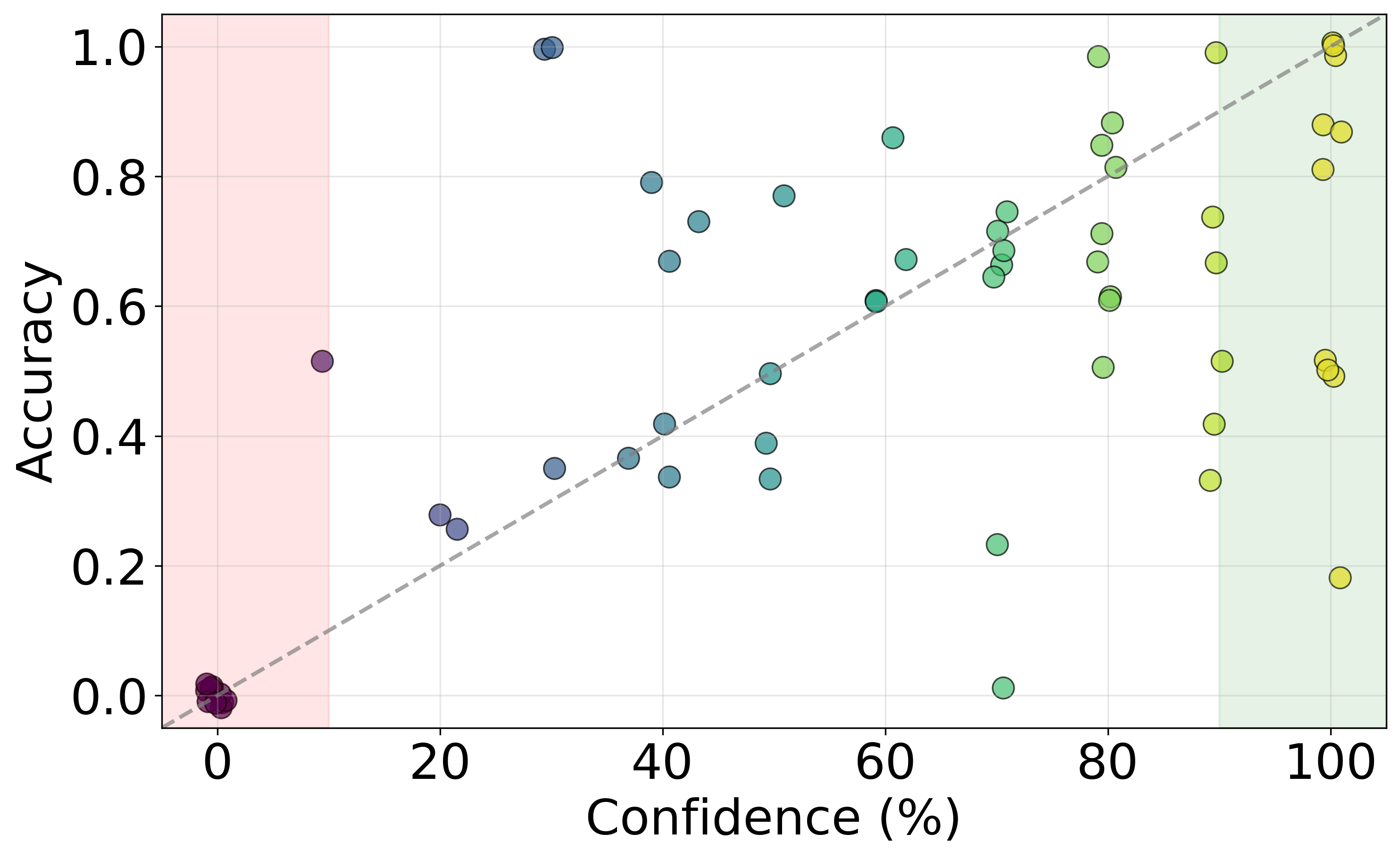}
    \caption{Gemini-2.5-Flash (MP setting)}
    \label{fig:mp_gemini_flash}
\end{figure}

\begin{figure}[ht!]
    \centering
    \includegraphics[width=0.85\linewidth]{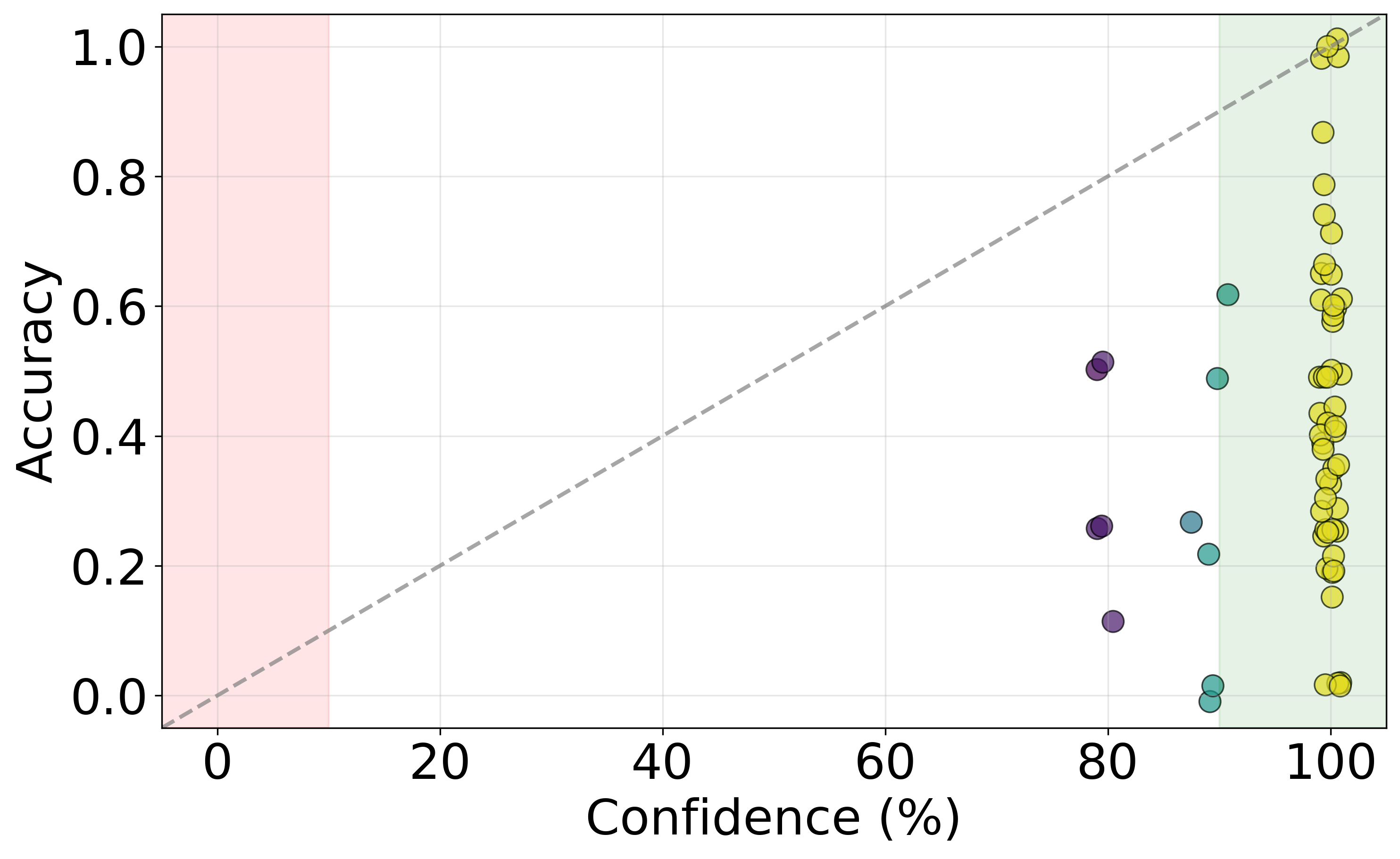}
    \caption{Llama-3.3-70B (MP setting)}
    \label{fig:mp_llama}
\end{figure}

\begin{figure}[ht!]
    \centering
    \includegraphics[width=0.85\linewidth]{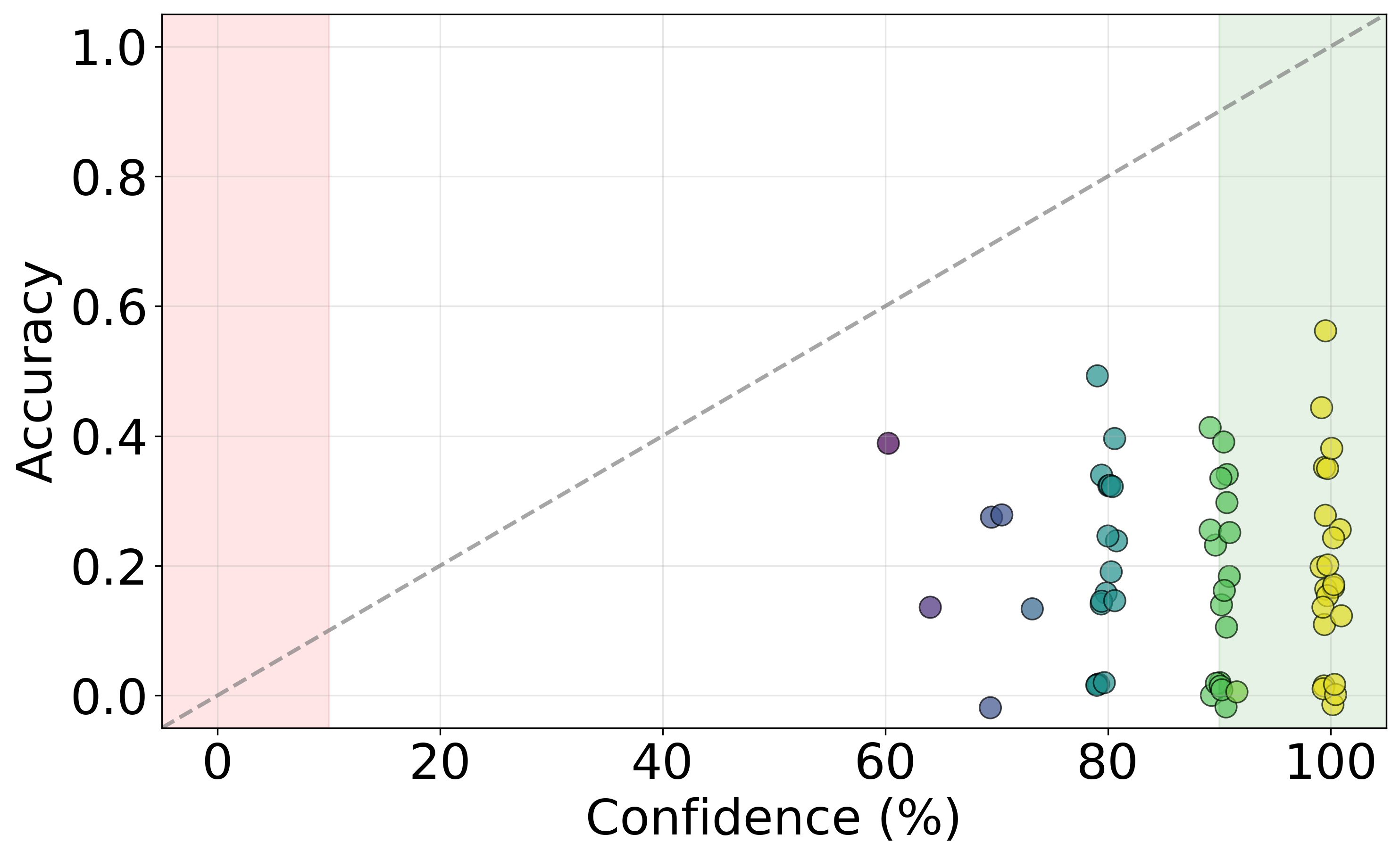}
    \caption{Mistral-Nemo (MP setting)}
    \label{fig:mp_mistral_nemo}
\end{figure}

\begin{figure}[ht!]
    \centering
    \includegraphics[width=0.85\linewidth]{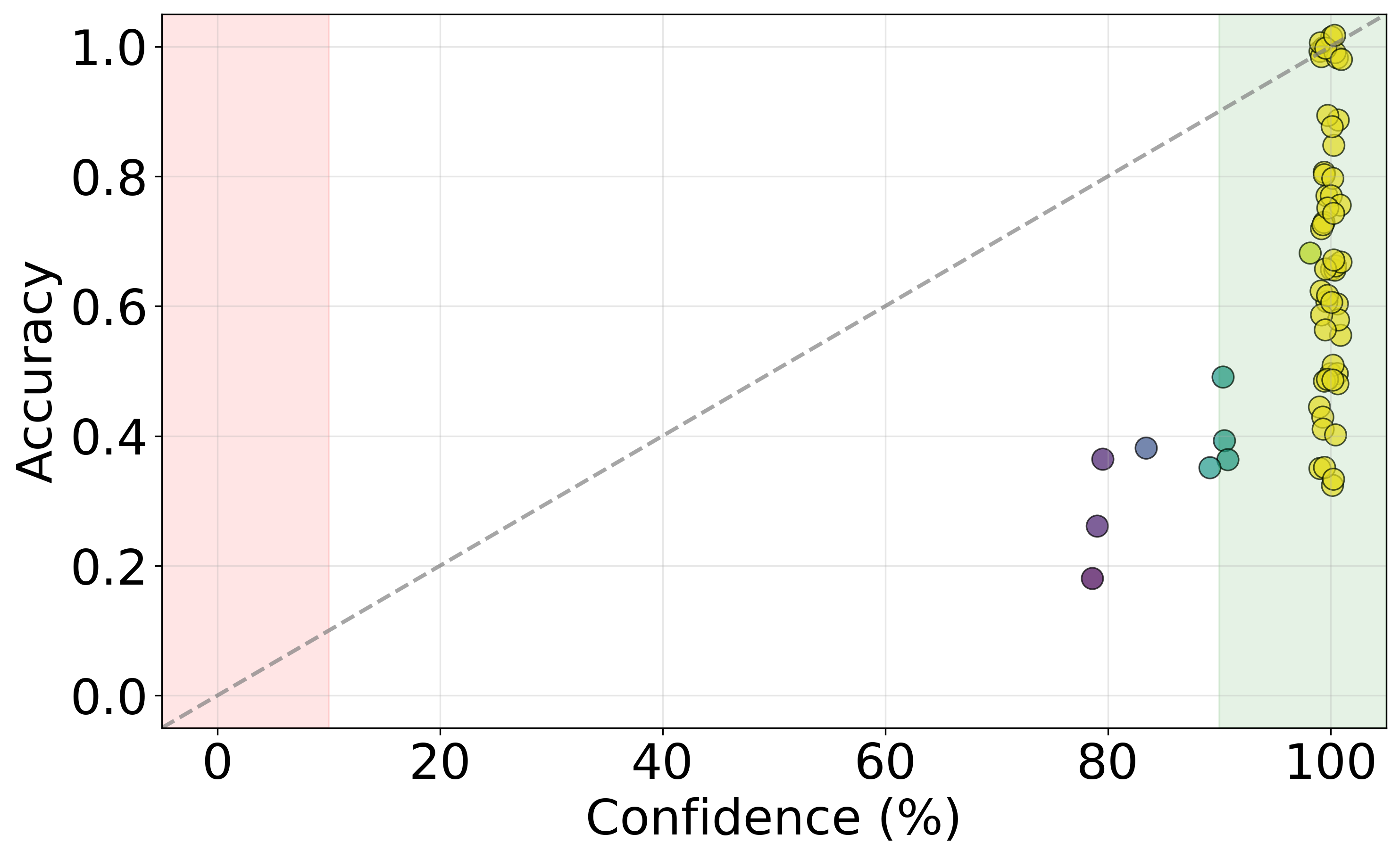}
    \caption{GPT-4.1 (MP setting)}
    \label{fig:mp_gpt41}
\end{figure}

\begin{figure}[ht!]
    \centering
    \includegraphics[width=0.85\linewidth]{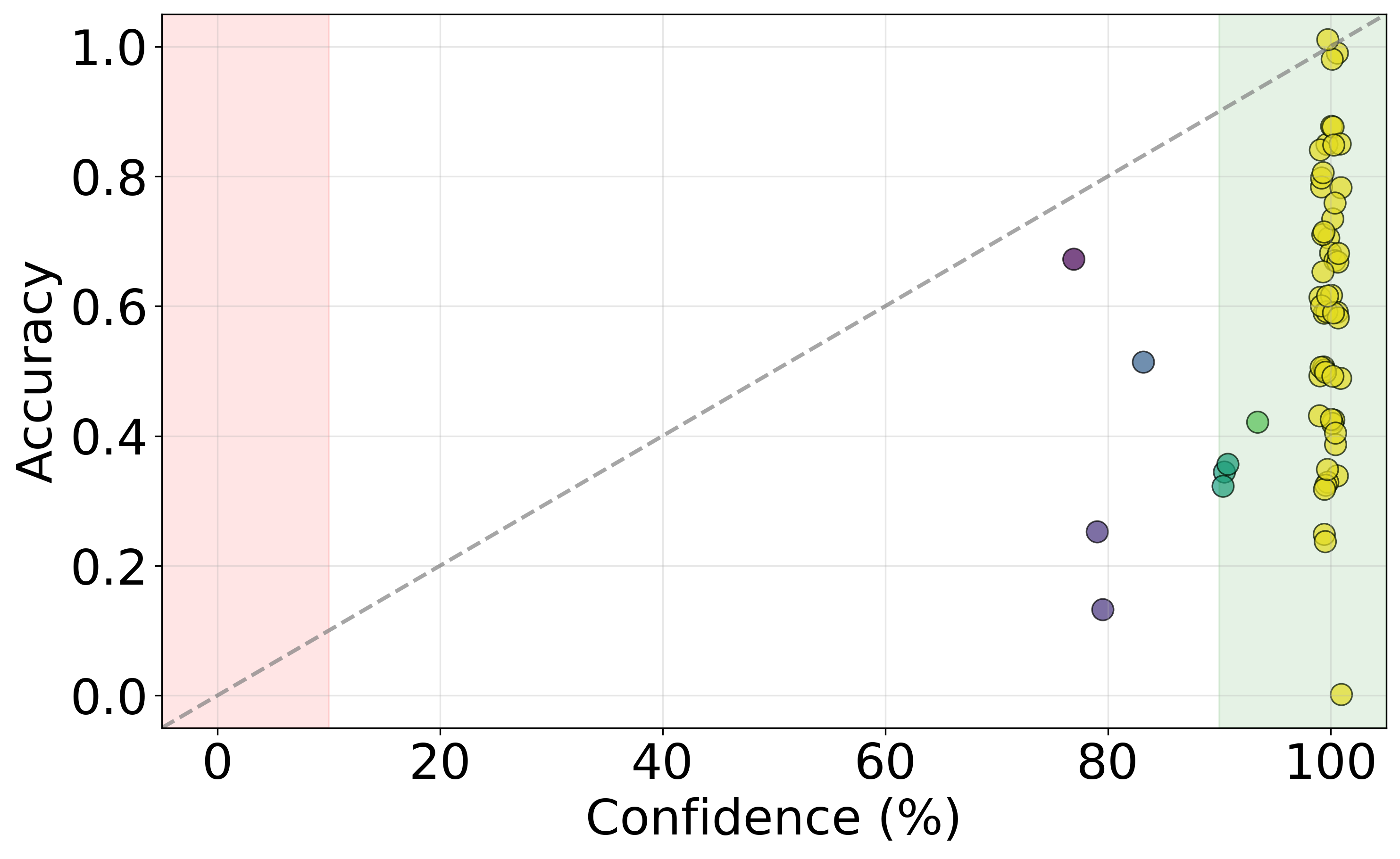}
    \caption{GPT-4.1-mini (MP setting)}
    \label{fig:mp2_gpt41_mini}
\end{figure}

\begin{figure}[ht!]
    \centering
    \includegraphics[width=0.85\linewidth]{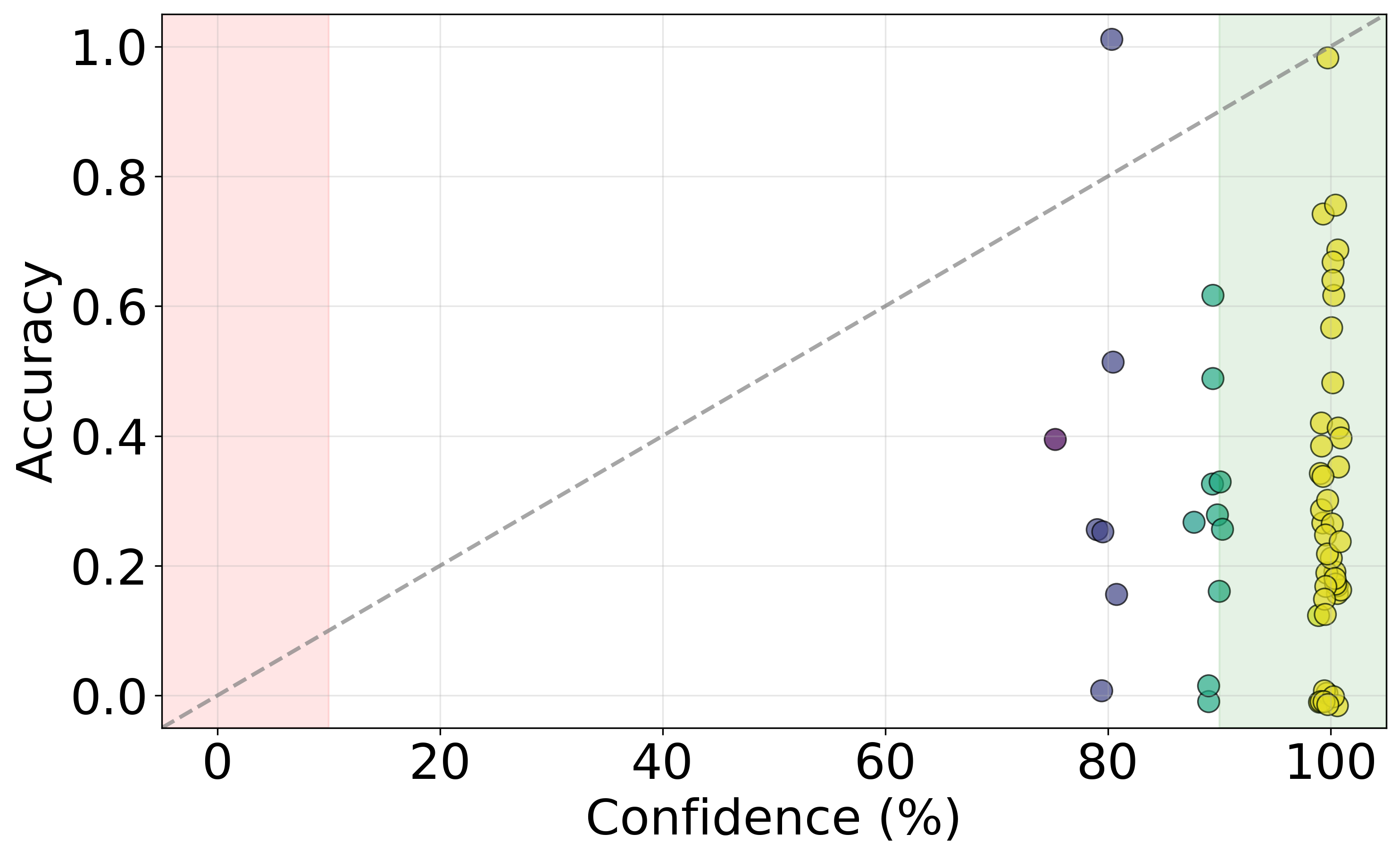}
    \caption{GPT-4.1-nano (MP setting)}
    \label{fig:mp2_gpt41_nano}
\end{figure}

\begin{figure}[ht!]
    \centering
    \includegraphics[width=0.85\linewidth]{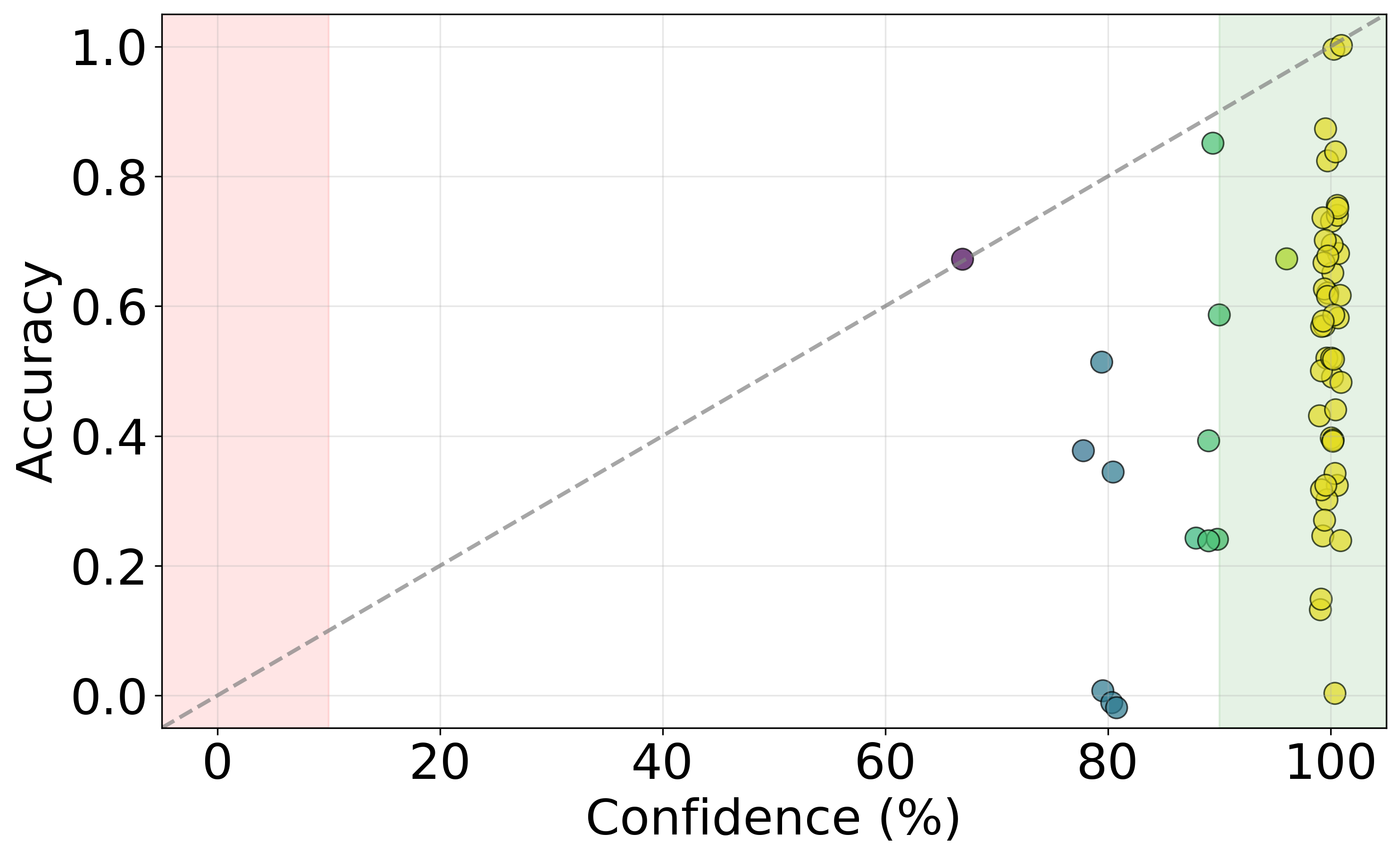}
    \caption{GPT-4o (MP setting)}
    \label{fig:mp2_gpt4o}
\end{figure}

\begin{figure}[ht!]
    \centering
    \includegraphics[width=0.85\linewidth]{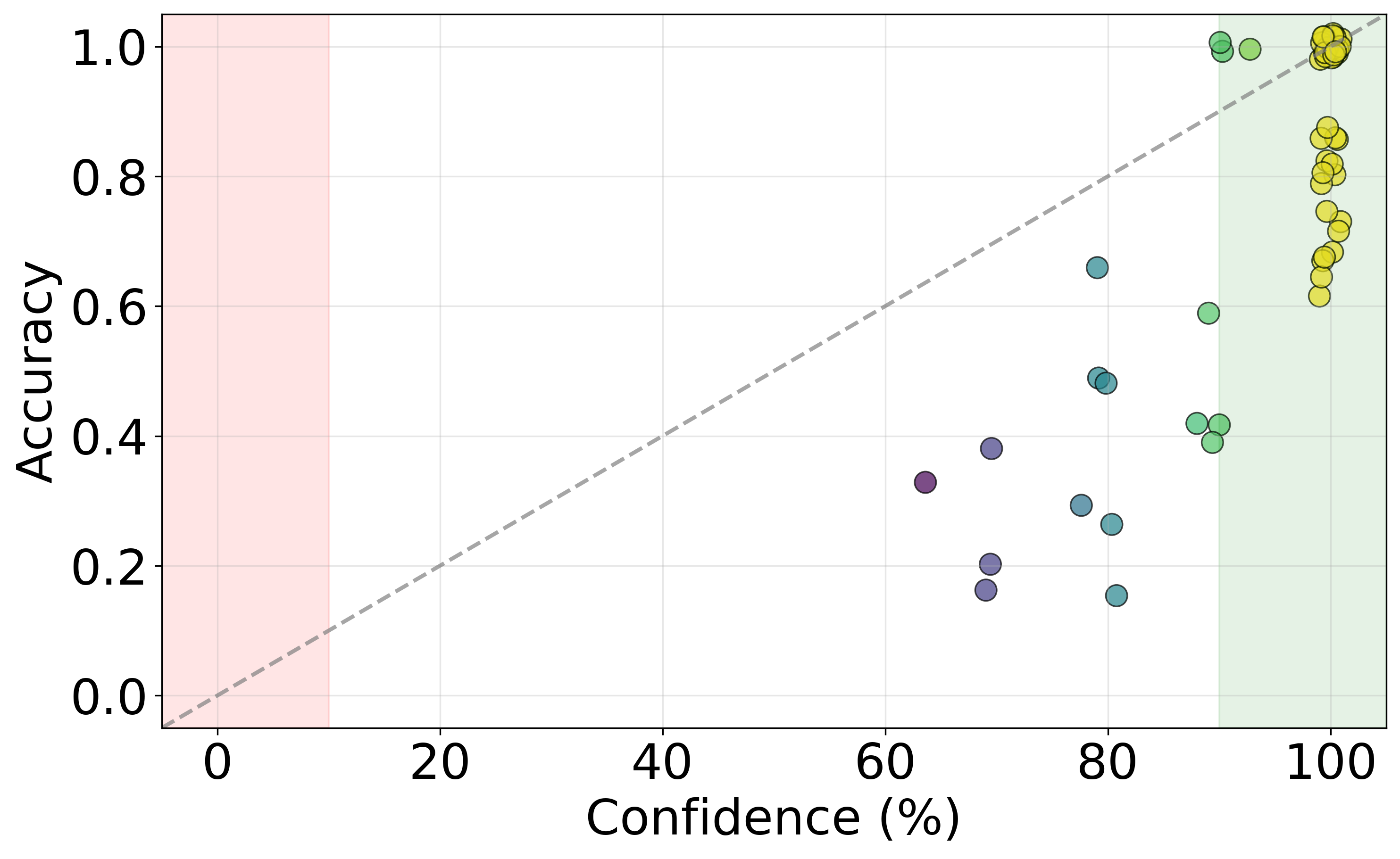}
    \caption{OpenAI-o3-mini (MP setting)}
    \label{fig:mp2_o3_mini}
\end{figure}

\begin{figure}[ht!]
    \centering
    \includegraphics[width=0.85\linewidth]{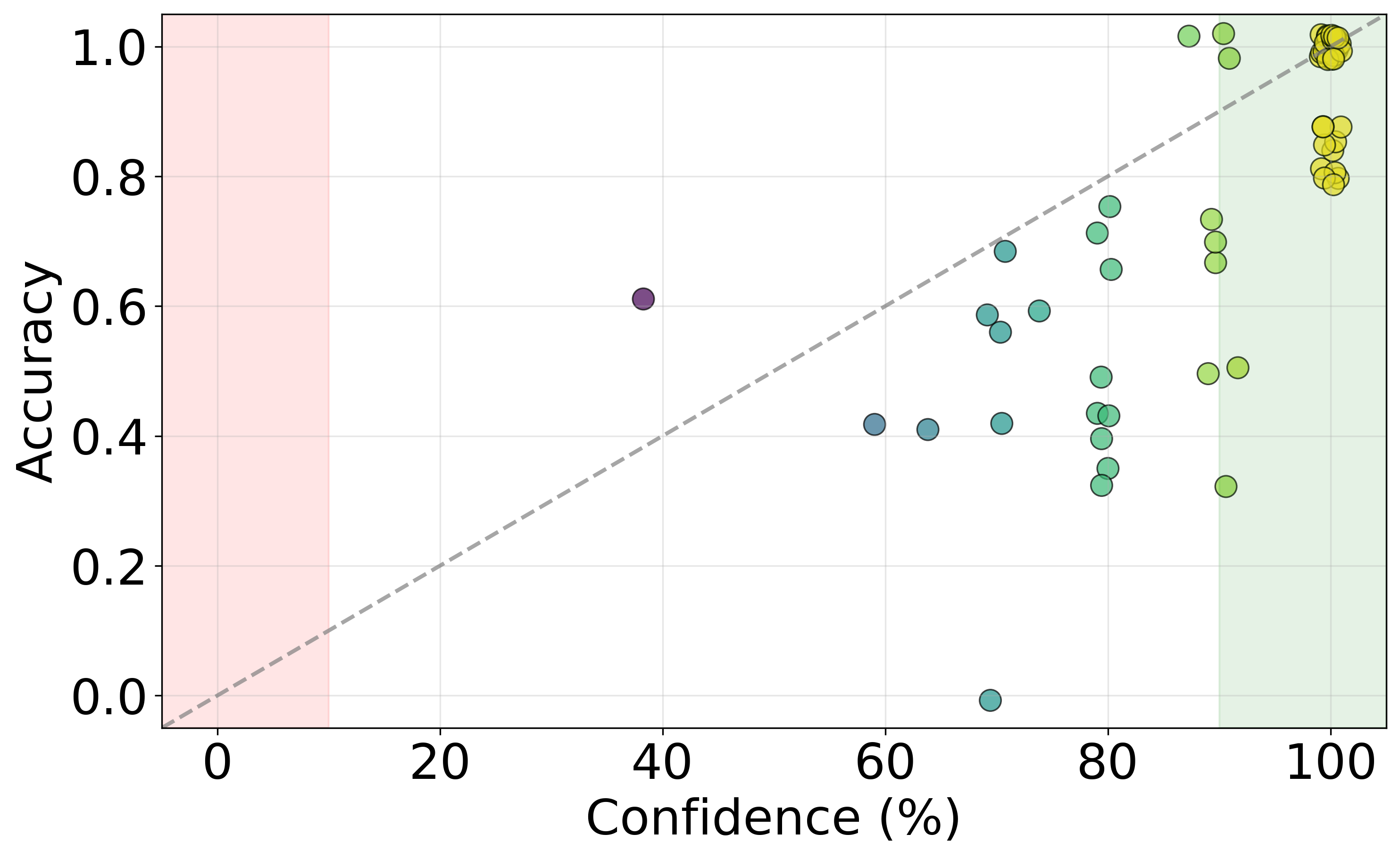}
    \caption{Qwen3-235B-A22B (MP setting)}
    \label{fig:mp2_qwen3}
\end{figure}

\begin{figure}[ht!]
    \centering
    \includegraphics[width=0.85\linewidth]{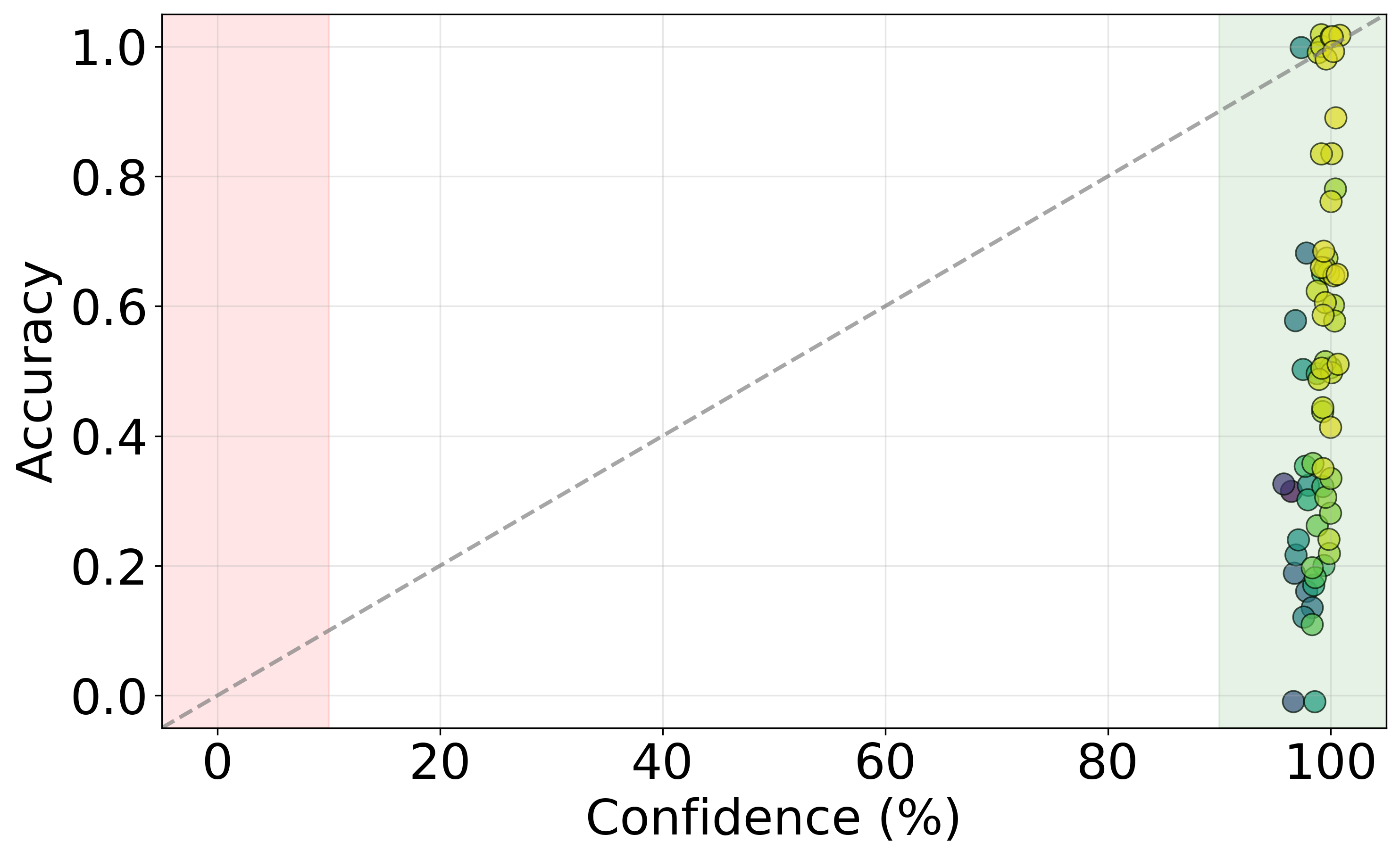}
    \caption{DeepSeek-V3-0324 (Logp setting)}
    \label{fig:log_deepseek_v3}
\end{figure}

\begin{figure}[ht!]
    \centering
    \includegraphics[width=0.85\linewidth]{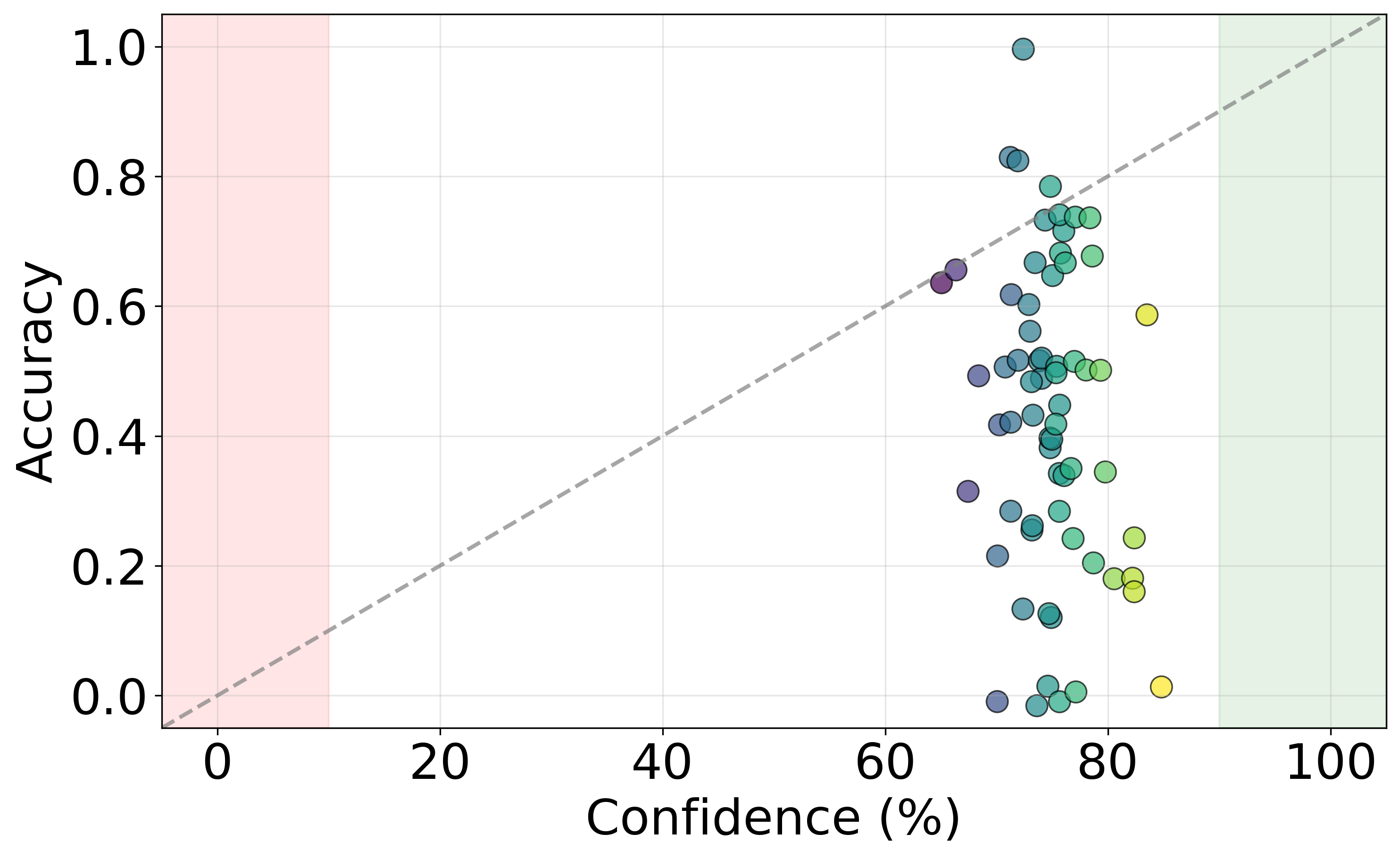}
    \caption{R1-Distill-Llama (Logp setting)}
    \label{fig:log_r1_distill_llama}
\end{figure}

\begin{figure}[ht!]
    \centering
    \includegraphics[width=0.85\linewidth]{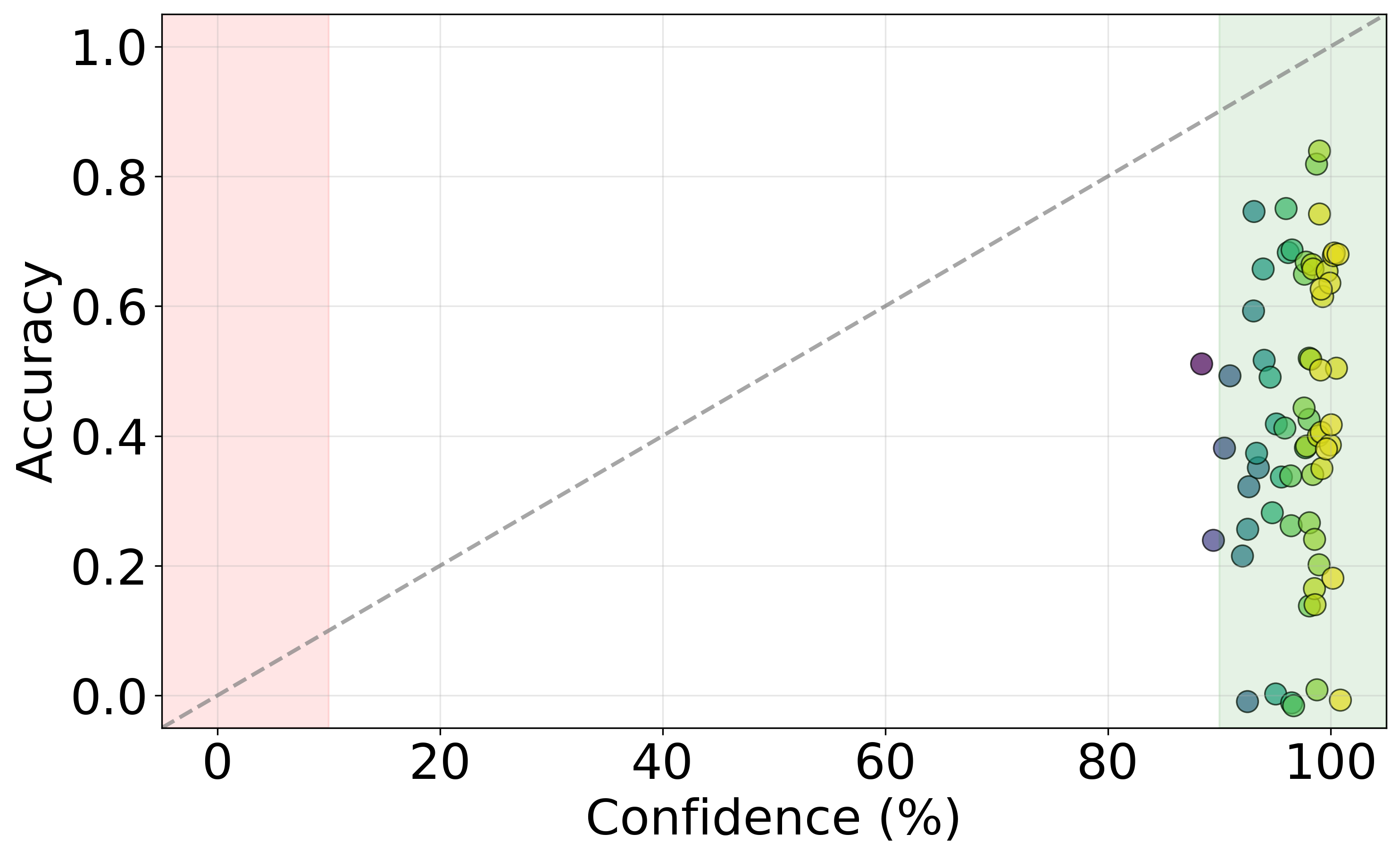}
    \caption{Llama-3.3-70B (Logp setting)}
    \label{fig:log_llama}
\end{figure}

\begin{figure}[ht!]
    \centering
    \includegraphics[width=0.85\linewidth]{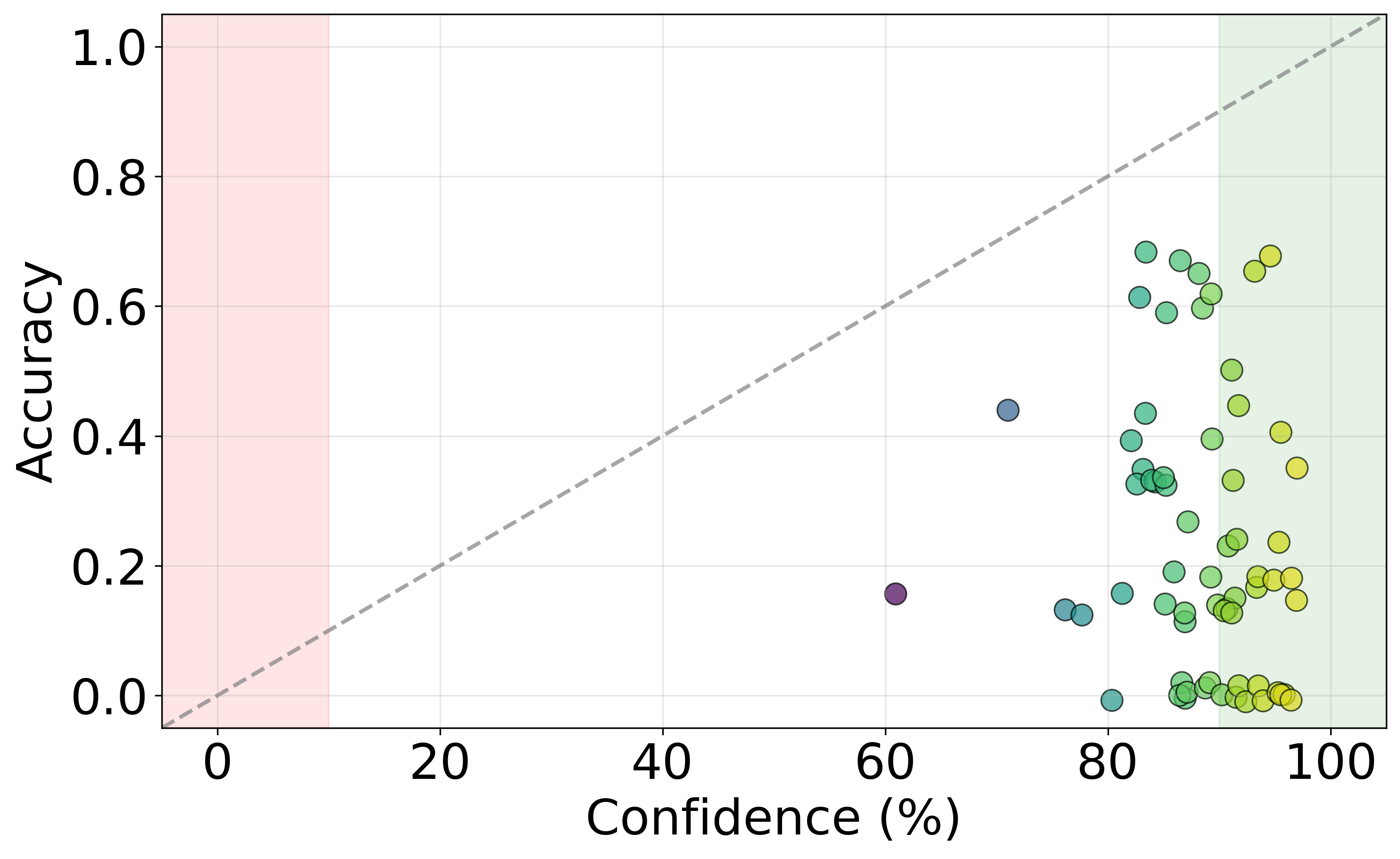}
    \caption{Mistral-Nemo (Logp setting)}
    \label{fig:log_mistral_nemo}
\end{figure}

\begin{figure}[ht!]
    \centering
    \includegraphics[width=0.85\linewidth]{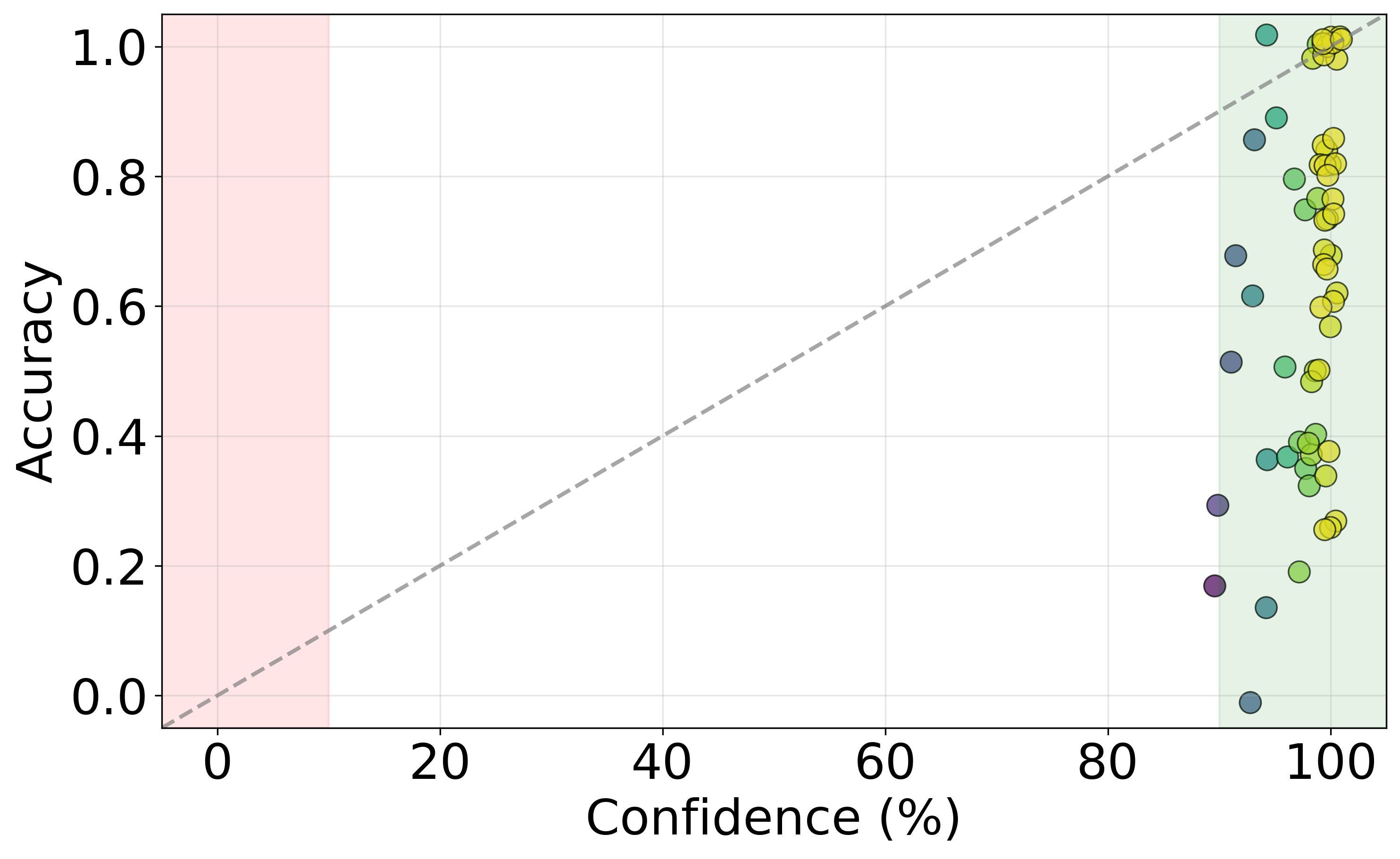}
    \caption{GPT-4.1 (Logp setting)}
    \label{fig:log_gpt41}
\end{figure}

\begin{figure}[ht!]
    \centering
    \includegraphics[width=0.85\linewidth]{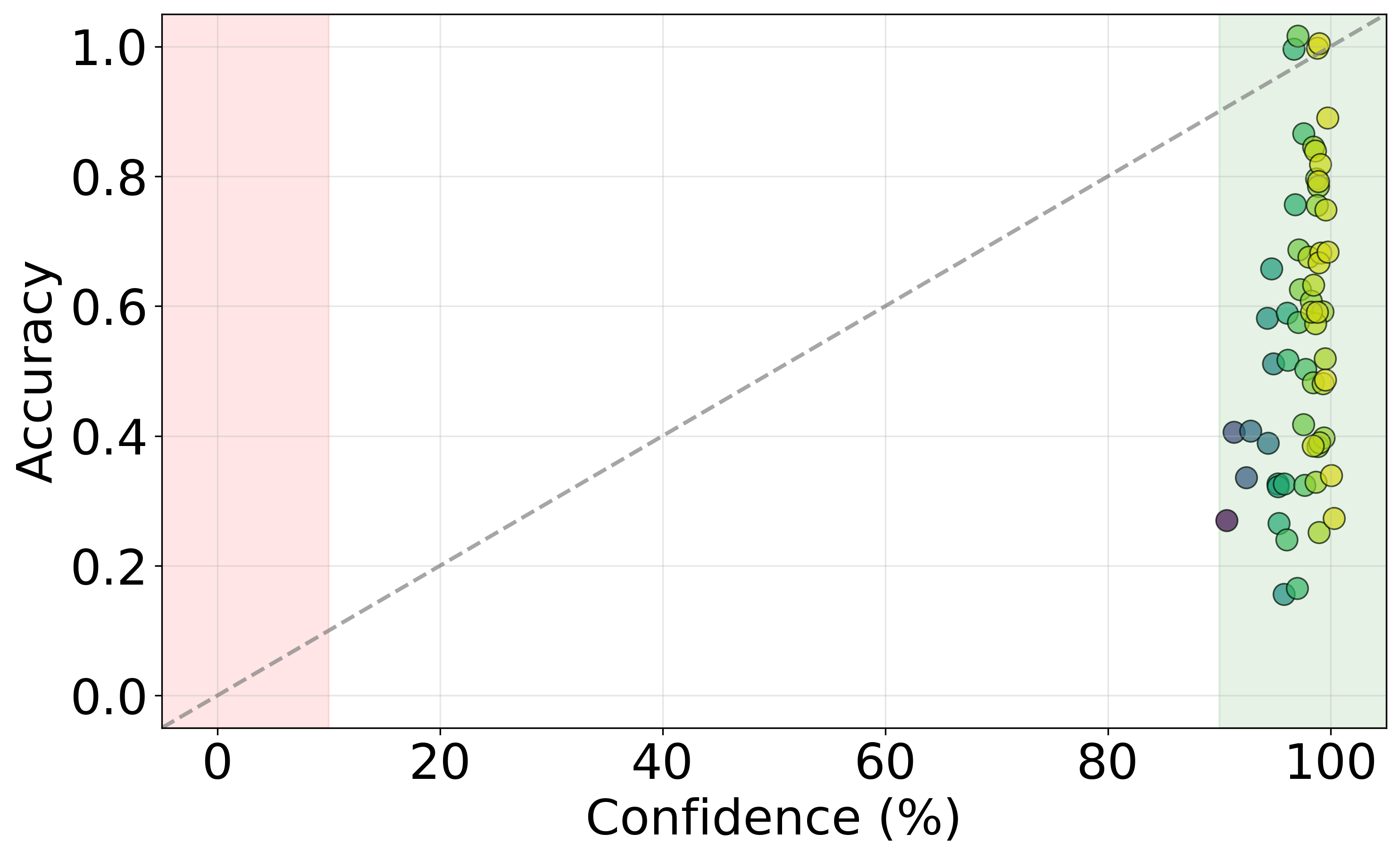}
    \caption{GPT-4.1-mini (Logp setting)}
    \label{fig:log_gpt41_mini}
\end{figure}

\begin{figure}[ht!]
    \centering
    \includegraphics[width=0.85\linewidth]{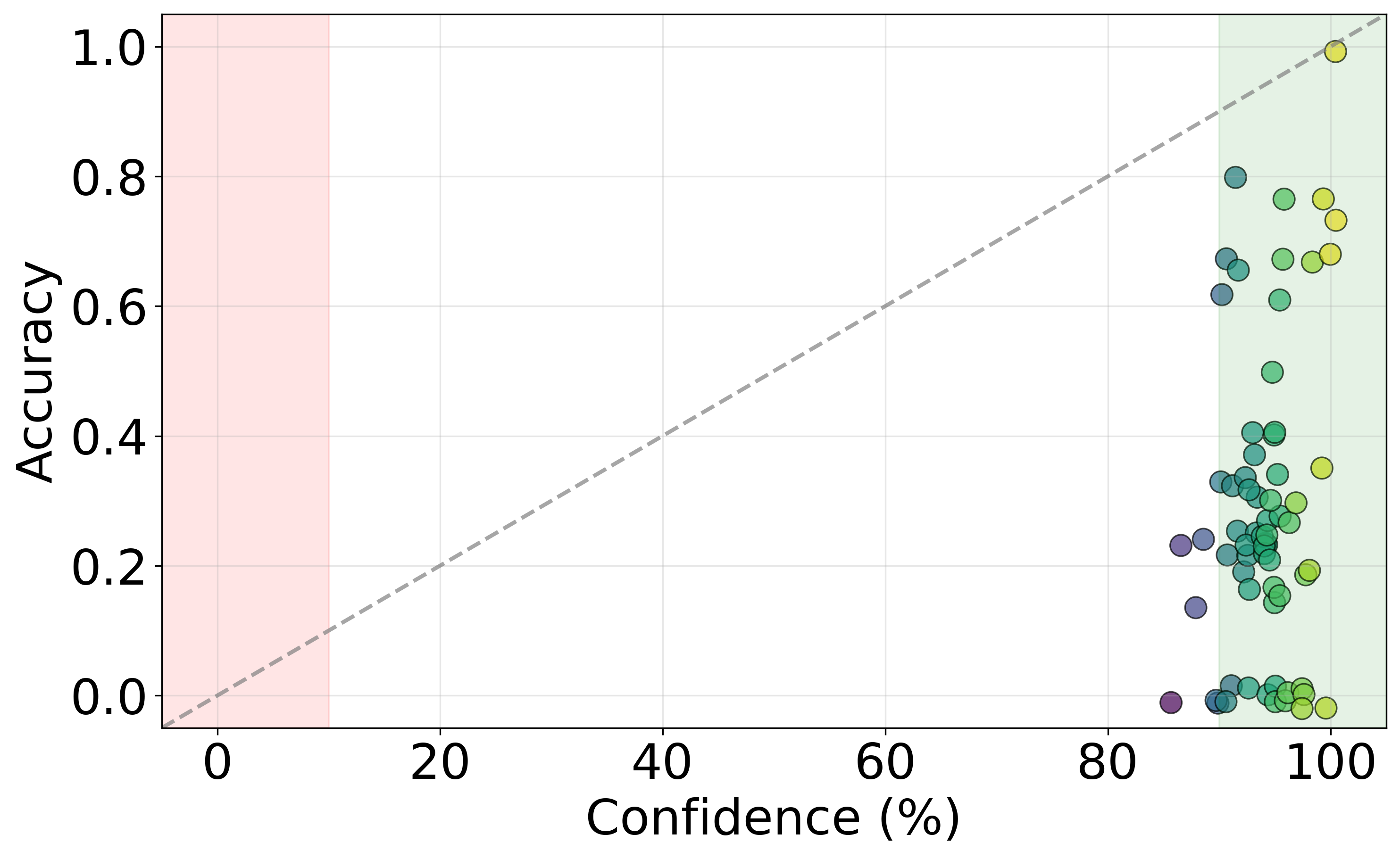}
    \caption{GPT-4.1-nano (Logp setting)}
    \label{fig:log2_gpt41_nano}
\end{figure}

\begin{figure}[ht!]
    \centering
    \includegraphics[width=0.85\linewidth]{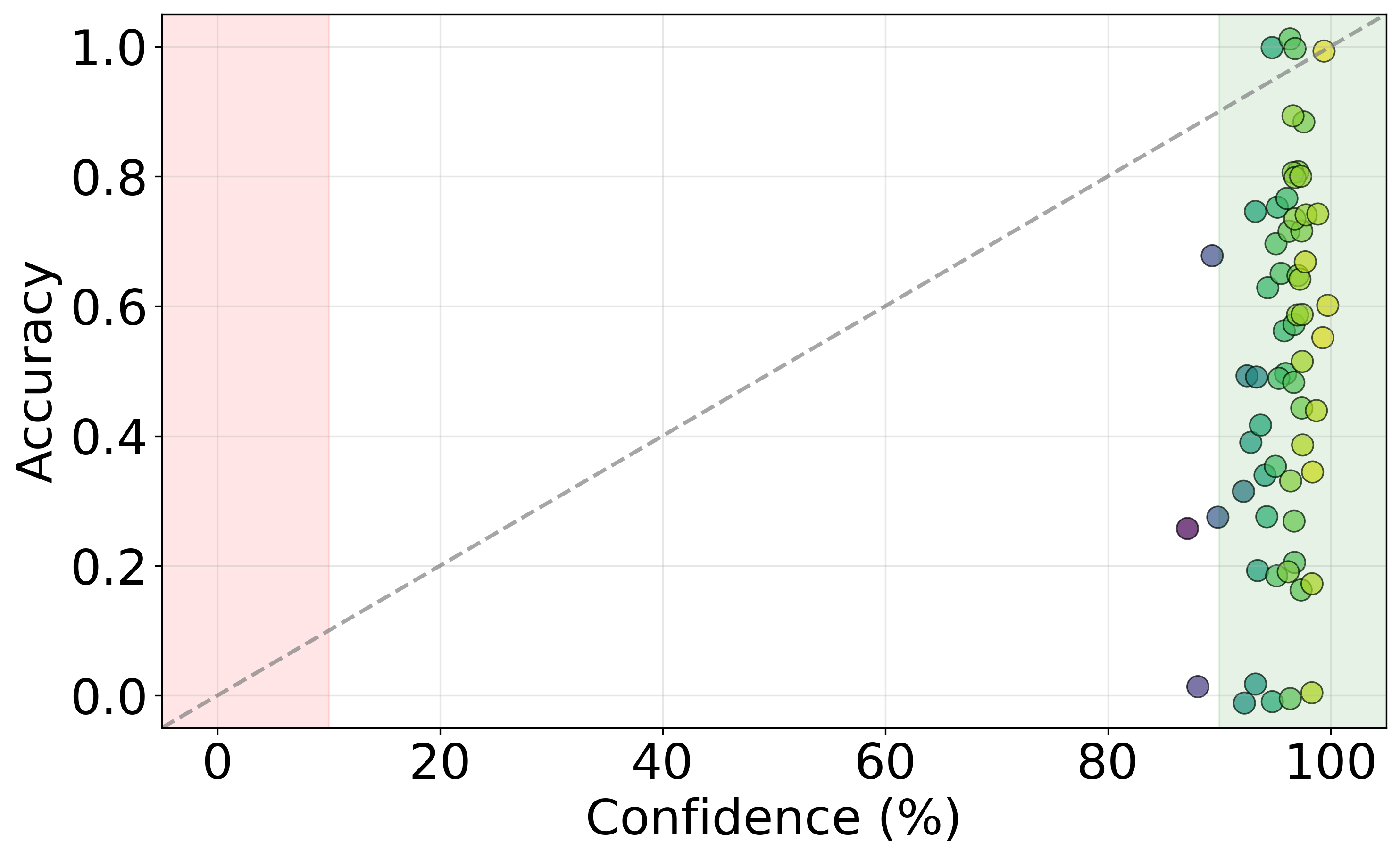}
    \caption{GPT-4o (Logp setting)}
    \label{fig:log2_gpt4o}
\end{figure}\begin{figure}[ht!]
    \centering
    \includegraphics[width=0.85\linewidth]{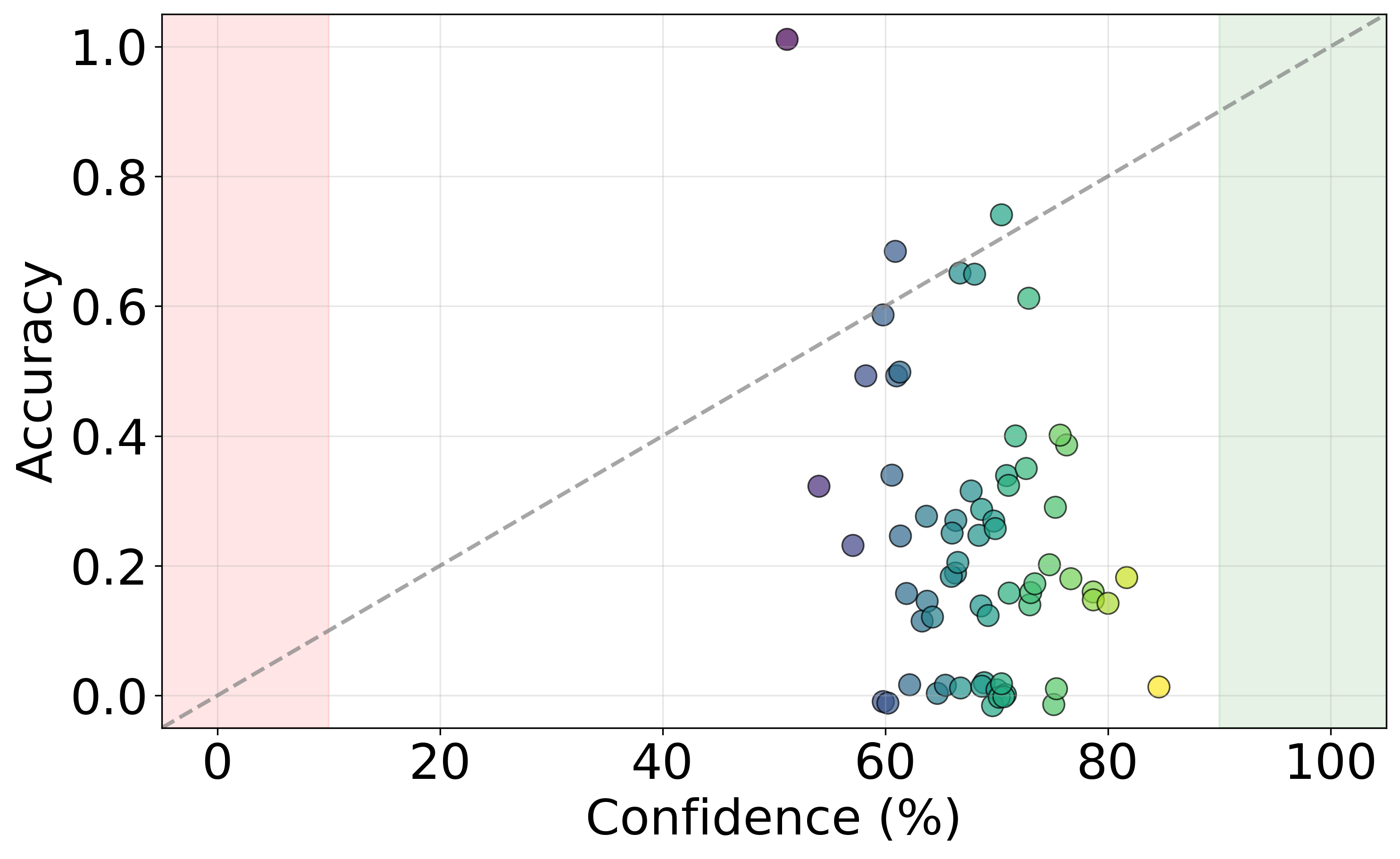}
    \caption{Qwen3-235B-A22B (Logp setting)}
    \label{fig:log2_qwen3}
\end{figure}\begin{figure}[ht!]
    \centering
    \includegraphics[width=0.85\linewidth]{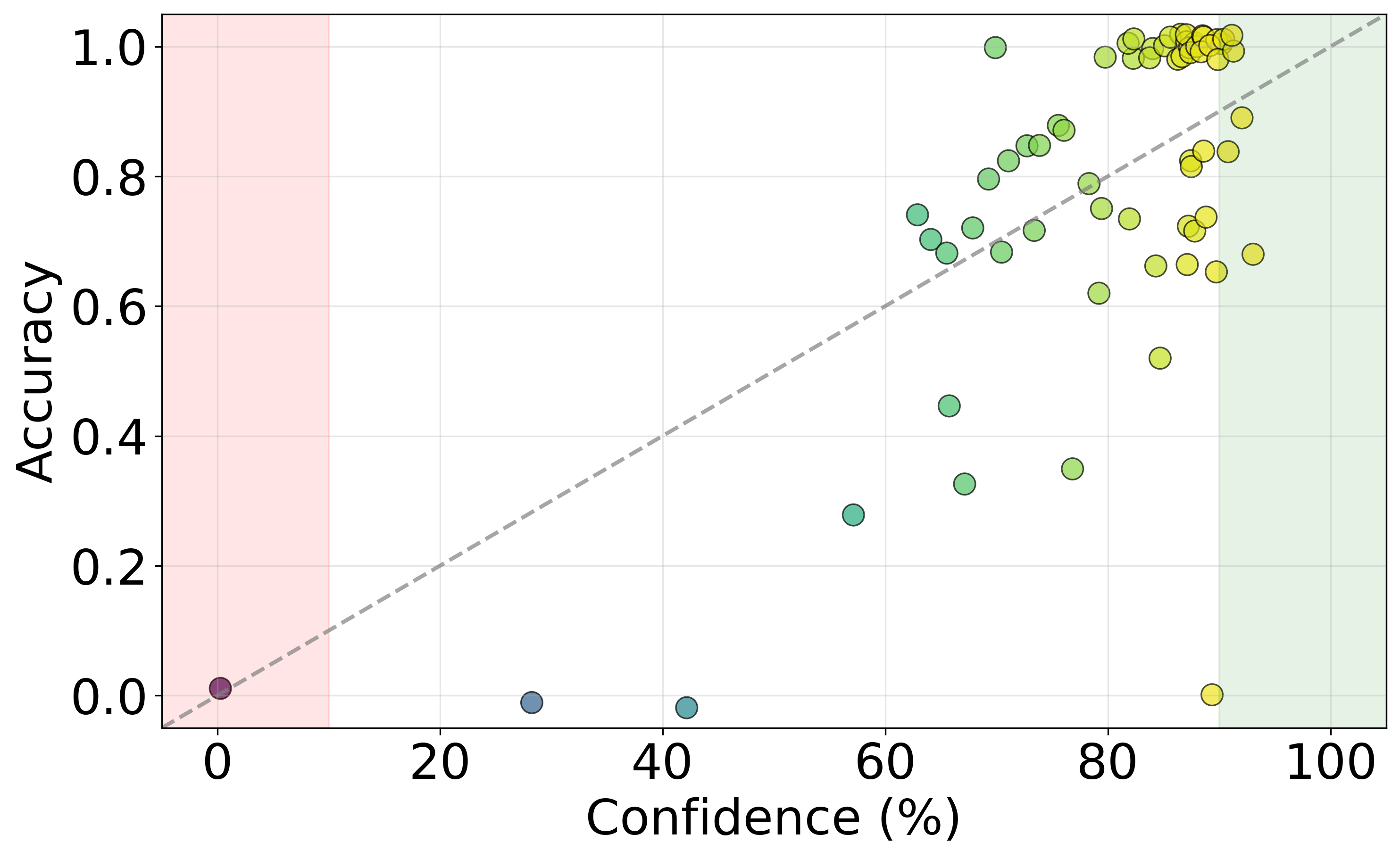}
    \caption{DeepSeek-R1-0528 (Logp setting)}
    \label{fig:log2_deepseek_r1}
\end{figure}

\begin{table*}[htbp]
\centering
\begin{tabular}{lccccccc}
\toprule
\textbf{Model} & \textbf{Acc} $\uparrow$ & \textbf{ECE} $\downarrow$ & \textbf{ACE} $\downarrow$ & \textbf{Brier Score} $\downarrow$ & \textbf{MCE} $\downarrow$ & \textbf{NLL} $\downarrow$ & \textbf{TH Score} $\uparrow$ \\ 
\midrule
DeepSeek-R1-0528 & \textbf{85.43} & \textbf{7.17} & \textbf{6.69} & \textbf{0.108} & \textbf{70.00} & \textbf{0.83} & \textbf{17.90} \\
Qwen3-235B-A22B & 78.86 & 13.00 & 12.04 & 0.151 & 70.00 & 0.98 & 16.85 \\
OpenAI-o3-mini & 76.00 & 18.49 & 18.77 & 0.184 & 43.91 & 1.84 & 17.08 \\
R1-Distill-Llama & 71.71 & 15.14 & 15.83 & 0.206 & 60.00 & 1.46 & 7.84 \\
R1-Distill-Qwen & 67.71 & 18.06 & 17.72 & 0.215 & 37.42 & 1.17 & 8.10 \\
Claude-Sonnet-4 & 64.29 & 34.51 & 34.48 & 0.340 & 65.00 & 6.12 & 9.17 \\
GPT-4.1 & 63.14 & 34.91 & 34.96 & 0.346 & 70.00 & 6.04 & 8.27 \\
Gemini-2.5-Flash & 52.57 & 14.43 & 14.77 & 0.220 & 43.33 & 1.44 & 7.40 \\
DeepSeek-V3-0324 & 50.57 & 47.89 & 47.96 & 0.479 & 70.00 & 9.02 & 0.79 \\
GPT-4o & 49.71 & 47.09 & 46.97 & 0.463 & 58.13 & 7.64 & 1.68 \\
GPT-4.1-mini & 56.00 & 42.31 & 42.18 & 0.419 & 65.00 & 7.52 & 4.35 \\
Llama-3.3-70B & 42.86 & 54.31 & 54.28 & 0.537 & 70.00 & 9.49 & -1.82 \\
GPT-4.1-nano & 28.29 & 67.43 & 67.41 & 0.663 & 71.26 & 10.86 & -6.76 \\
Mistral-Nemo & 19.43 & 68.89 & 68.93 & 0.643 & 78.69 & 6.69 & -4.35 \\
\bottomrule
\end{tabular}
\caption{Performance Comparison of Different LLMs under MP Setting}
\label{table:mp_results}
\end{table*}

\begin{table*}[htbp]
\centering
\begin{tabular}{lccccccc}
\toprule
\textbf{Model} & \textbf{Acc} $\uparrow$ & \textbf{ECE} $\downarrow$ & \textbf{ACE} $\downarrow$ & \textbf{Brier Score} $\downarrow$ & \textbf{MCE} $\downarrow$ & \textbf{NLL} $\downarrow$ & \textbf{TH Score} $\uparrow$ \\ 
\midrule
DeepSeek-R1-0528 & \textbf{78.29} & \textbf{6.84} & \textbf{6.62} & \textbf{0.1298} & \textbf{46.15} & \textbf{0.4211} & 2.96 \\
GPT-4.1 & 63.43 & 34.46 & 34.43 & 0.3462 & 63.80 & 1.7287 & \textbf{7.36} \\
GPT-4.1-mini & 55.14 & 42.56 & 42.53 & 0.4253 & 58.32 & 1.7946 & 2.56 \\
GPT-4o & 50.86 & 45.05 & 45.06 & 0.4493 & 61.19 & 1.6238 & 0.79 \\
DeepSeek-V3-0324 & 48.29 & 50.76 & 50.68 & 0.5044 & 50.76 & 2.4714 & -0.85 \\
Llama-3.3-70B & 43.43 & 53.53 & 53.55 & 0.5318 & 54.04 & 2.3400 & -3.25 \\
GPT-4.1-nano & 28.00 & 66.05 & 66.16 & 0.6349 & 82.98 & 2.1206 & -8.70 \\
Mistral-Nemo & 23.43 & 64.63 & 64.60 & 0.6051 & 79.59 & 1.7887 & -5.89 \\
\bottomrule
\end{tabular}
\caption{Performance Comparison of Different LLMs under LogP Setting}
\label{table:logp_results}
\end{table*}

\end{document}